\documentclass[10pt,twocolumn,letterpaper]{article}
\usepackage{caption, subcaption}
\usepackage{iccv}
\usepackage{times}
\usepackage{epsfig}
\usepackage{graphicx}
\usepackage{amsmath}
\usepackage{amssymb}
\usepackage[accsupp]{axessibility}  

\usepackage{algorithm}
\usepackage{algorithmic}
\usepackage{listings}

\usepackage[table]{xcolor}

\usepackage{enumitem}
\usepackage{cancel}
\usepackage{tablefootnote}
\usepackage{float}
\usepackage{makecell}
\usepackage{booktabs}
\usepackage{multirow}
\usepackage{ulem}
\usepackage{pifont}
\newcommand{\cmark}{\ding{51}}%
\newcommand{\xmark}{\ding{55}}%

\newcolumntype{?}[1]{!{\vrule width #1}}

\usepackage[pagebackref=true,breaklinks=true,letterpaper=true,colorlinks,bookmarks=false]{hyperref}

\iccvfinalcopy 

\usepackage[capitalize]{cleveref}
\crefname{section}{Sec.}{Secs.}
\Crefname{section}{Section}{Sections}
\Crefname{table}{Table}{Tables}
\crefname{table}{Tab.}{Tabs.}

\ificcvfinal
\begin{document}

\title{
AlignDet: Aligning Pre-training and Fine-tuning in Object Detection
}

\author{Ming Li$^{1}$$^{\star}$ \quad 
Jie Wu$^{1 \star}$$^\dagger$ \quad
Xionghui Wang$^{1}$ \quad
Chen Chen$^{2}$$^\dagger$ \\
Jie Qin$^{1}$ \quad
Xuefeng Xiao$^{1}$ \quad
Rui Wang$^{1}$ \quad
Min Zheng$^{1}$ \quad
Xin Pan$^{1}$ \\
$^{1}$ByteDance Inc \
$^{2}$Center for Research in Computer Vision, University of Central Florida
}


\twocolumn[{%
\renewcommand\twocolumn[1][]{#1}%
\maketitle
\begin{center}
    \vspace{-20pt}
    \includegraphics[width=0.99\linewidth,height=0.35\linewidth]{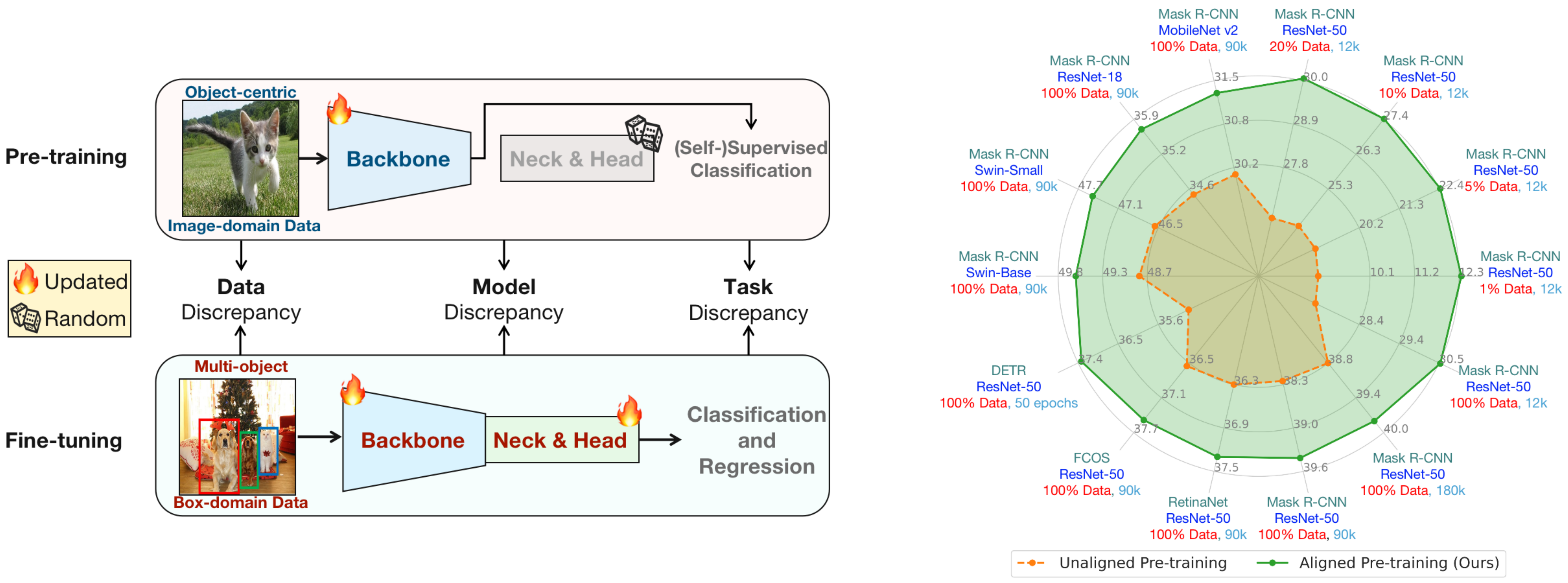}
\captionof{figure}{Illustration of the \textbf{data}, \textbf{model}, and \textbf{task} discrepancies between the pre-training and fine-tuning stages in object detection. In mostly pre-training phase, only the backbone is updated via classification task supervision on object-centric datasets such as ImageNet. However, the whole detector is fine-tuned in the multi-objects-based datasets, supervising via classification and regression tasks. By bridging these discrepancies with self-supervised pre-training, AlignDet achieves significant improvements across various \textcolor{teal}{detection algorithms}, \textcolor{blue}{model backbones}, \textcolor{red}{data settings}, and \textcolor{cyan}{training schedules} on COCO. \textcolor{magenta}{\href{https://liming-ai.github.io/AlignDet}{Project Page: https://liming-ai.github.io/AlignDet}}.
}
    \vspace{3pt}
    \label{fig:motivation}
    \end{center}%
}]

\renewcommand{\thefootnote}{}
\footnotetext{$^\star$Equal contribution. $^\dagger$Corresponding author.
}

\begin{abstract}
\vspace{-0.4cm}
The paradigm of large-scale pre-training followed by downstream fine-tuning has been widely employed in various object detection algorithms. 
In this paper, we reveal discrepancies in \textbf{data}, \textbf{model}, and \textbf{task} between the pre-training and fine-tuning procedure in existing practices, which implicitly limit the detector's performance, generalization ability, and convergence speed.
To this end, we propose AlignDet, a unified pre-training framework that can be adapted to various existing detectors to alleviate the discrepancies. AlignDet decouples the pre-training process into two stages, i.e., image-domain and box-domain pre-training. 
The image-domain pre-training optimizes the detection backbone to capture holistic visual abstraction, and box-domain pre-training learns instance-level semantics and task-aware concepts to initialize the parts out of the backbone.
By incorporating the self-supervised pre-trained backbones, we can pre-train all modules for various detectors in an unsupervised paradigm.
As depicted in Figure \ref{fig:motivation}, extensive experiments demonstrate that AlignDet can achieve significant improvements across diverse protocols, such as \textcolor{teal}{detection algorithm}, \textcolor{blue}{model backbone}, \textcolor{red}{data setting}, and \textcolor{cyan}{training schedule}. 
For example, AlignDet improves FCOS by \textbf{5.3} mAP, RetinaNet by  \textbf{2.1} mAP, Faster R-CNN by \textbf{3.3} mAP, and DETR by \textbf{2.3} mAP under fewer epochs.

\vspace{-0.35cm}

\end{abstract}

\begin{table*}[]
\centering
\resizebox{\linewidth}{!}{%
    \begin{tabular}{c?{0.4mm}ccccccc?{0.4mm}ccc}
    \multirow{3}{*}{\textbf{Method}} &
      \multicolumn{7}{c?{0.4mm}}{\textbf{Discrepancy}} &
      \multicolumn{3}{c}{\textbf{Generalization}} \\ \cline{2-11} 
     &
      \multicolumn{2}{c|}{\textbf{Data}} &
      \multicolumn{3}{c|}{\textbf{Model}} &
      \multicolumn{2}{c?{0.4mm}}{\textbf{Task}} &
      \multicolumn{1}{c|}{\multirow{2}{*}{\textbf{Anchor-based}}} &
      \multicolumn{1}{c|}{\multirow{2}{*}{\textbf{Point-based}}} &
      \multirow{2}{*}{\textbf{Query-based}} \\ \cline{2-8}
     &
      \multicolumn{1}{c|}{\textbf{Object-centric}} &
      \multicolumn{1}{c|}{\textbf{Multi-object}} &
      \multicolumn{1}{c|}{\textbf{Backbone}} &
      \multicolumn{1}{c|}{\textbf{Neck}} &
      \multicolumn{1}{c|}{\textbf{Head}} &
      \multicolumn{1}{c|}{\textbf{Classification}} &
      \textbf{Regression} &
      \multicolumn{1}{c|}{} &
      \multicolumn{1}{c|}{} &
       \\ \hline
    Supervised Backbone &
      \multicolumn{1}{c|}{\cmark} &
      \multicolumn{1}{c|}{\xmark} &
      \multicolumn{1}{c|}{\cmark} &
      \multicolumn{1}{c|}{\xmark} &
      \multicolumn{1}{c|}{\xmark} &
      \multicolumn{1}{c|}{\cmark} &
      \xmark &
      \multicolumn{1}{c|}{\cmark} &
      \multicolumn{1}{c|}{\cmark} &
      \cmark \\ \hline
    MoCo~\cite{moco}, SwAV~\cite{swav} &
      \multicolumn{1}{c|}{\cmark} &
      \multicolumn{1}{c|}{\xmark} &
      \multicolumn{1}{c|}{\cmark} &
      \multicolumn{1}{c|}{\xmark} &
      \multicolumn{1}{c|}{\xmark} &
      \multicolumn{1}{c|}{\cmark} &
      \xmark &
      \multicolumn{1}{c|}{\cmark} &
      \multicolumn{1}{c|}{\cmark} &
      \cmark \\ \hline
    DenseCL~\cite{densecl}, SelfEMD~\cite{self_emd}&
      \multicolumn{1}{c|}{\cmark} &
      \multicolumn{1}{c|}{\cmark} &
      \multicolumn{1}{c|}{\cmark} &
      \multicolumn{1}{c|}{\xmark} &
      \multicolumn{1}{c|}{\xmark} &
      \multicolumn{1}{c|}{\cmark} &
      \xmark &
      \multicolumn{1}{c|}{\cmark} &
      \multicolumn{1}{c|}{\cmark} &
      \cmark \\ \hline
    PixPro~\cite{pixpro}, InsLoc~\cite{insloc}, SoCo~\cite{soco} &
      \multicolumn{1}{c|}{\cmark} &
      \multicolumn{1}{c|}{\cmark} &
      \multicolumn{1}{c|}{\cmark} &
      \multicolumn{1}{c|}{\cmark} &
      \multicolumn{1}{c|}{\xmark} &
      \multicolumn{1}{c|}{\cmark} &
      \xmark &
      \multicolumn{1}{c|}{\cmark} &
      \multicolumn{1}{c|}{\cmark} &
      \xmark \\ \hline
    UP-DETR~\cite{up_detr}, DETReg~\cite{detreg}&
      \multicolumn{1}{c|}{\cmark} &
      \multicolumn{1}{c|}{\cmark} &
      \multicolumn{1}{c|}{\cmark} &
      \multicolumn{1}{c|}{\cmark} &
      \multicolumn{1}{c|}{\cmark} &
      \multicolumn{1}{c|}{\xmark} &
      \cmark &
      \multicolumn{1}{c|}{\xmark} &
      \multicolumn{1}{c|}{\xmark} &
      \cmark \\ \hline
    \rowcolor{gray!10}Ours &
      \multicolumn{1}{c|}{\cmark} &
      \multicolumn{1}{c|}{\cmark} &
      \multicolumn{1}{c|}{\cmark} &
      \multicolumn{1}{c|}{\cmark} &
      \multicolumn{1}{c|}{\cmark} &
      \multicolumn{1}{c|}{\cmark} &
      \cmark &
      \multicolumn{1}{c|}{\cmark} &
      \multicolumn{1}{c|}{\cmark} &
      \cmark
      
    \end{tabular}
}
\vspace{-0.2cm}
\caption{We compare with other unsupervised pre-training algorithms in terms of solving the discrepancies and the generalization ability. Taking task discrepancy as an example, previous methods can only introduce the pretext task of classification or regression in the pre-training stage,  while AlignDet can build both tasks to learn the semantic and positional information of the objects.}
\label{tab:comparison_discrepancy}
\end{table*}

\section{Introduction}
\label{sec:intro}
In recent years, there has been significant progress in large-scale pre-training and fine-tuning optimization paradigms in computer vision.
A series of pre-training algorithms have been designed to learn domain-sensitive or task-aware concepts to boost downstream performance~\cite{fast_rcnn,moco,beit}.
As for object detection, current approaches generally leverage ImageNet~\cite{imagenet} to pre-train the backbone with classification-oriented supervision.
However, compared to the detection-oriented fine-tuning  process, this pre-training paradigm exhibits  three discrepancies, as shown in Figure~\ref{fig:motivation}:

\vspace{-0.1cm}

\begin{itemize}[leftmargin=*] 
\item \textit{Data}: most pre-training methods are conducted on single object-centric datasets, like ImageNet. However, the detection datasets, \emph{e.g.}, COCO~\cite{coco}, usually consist of multiple objects with different scales and locations. The differences in data characteristics and domain can cause the pre-training to deviate from the downstream task.

\item  \textit{Model}: current pre-training algorithms mainly focus on partial modules (such as the backbone) within the model due to the diversity and complexity of detectors. Some key detection components (such as the RPN and regression head) remain random initialization.

\item  \textit{Task}: existing pre-training approaches only regard the classification task as the pretext task, failing to capture object-aware positional context, including proposal generation, target assignment, and box regression.
\end{itemize}

These discrepancies potentially bring limited results, poor generalization, and slow convergence speed~\cite{selfsup_helps_selfsup,obj365}.

\begin{figure}[t]\centering
    \includegraphics[width=1.0\linewidth]{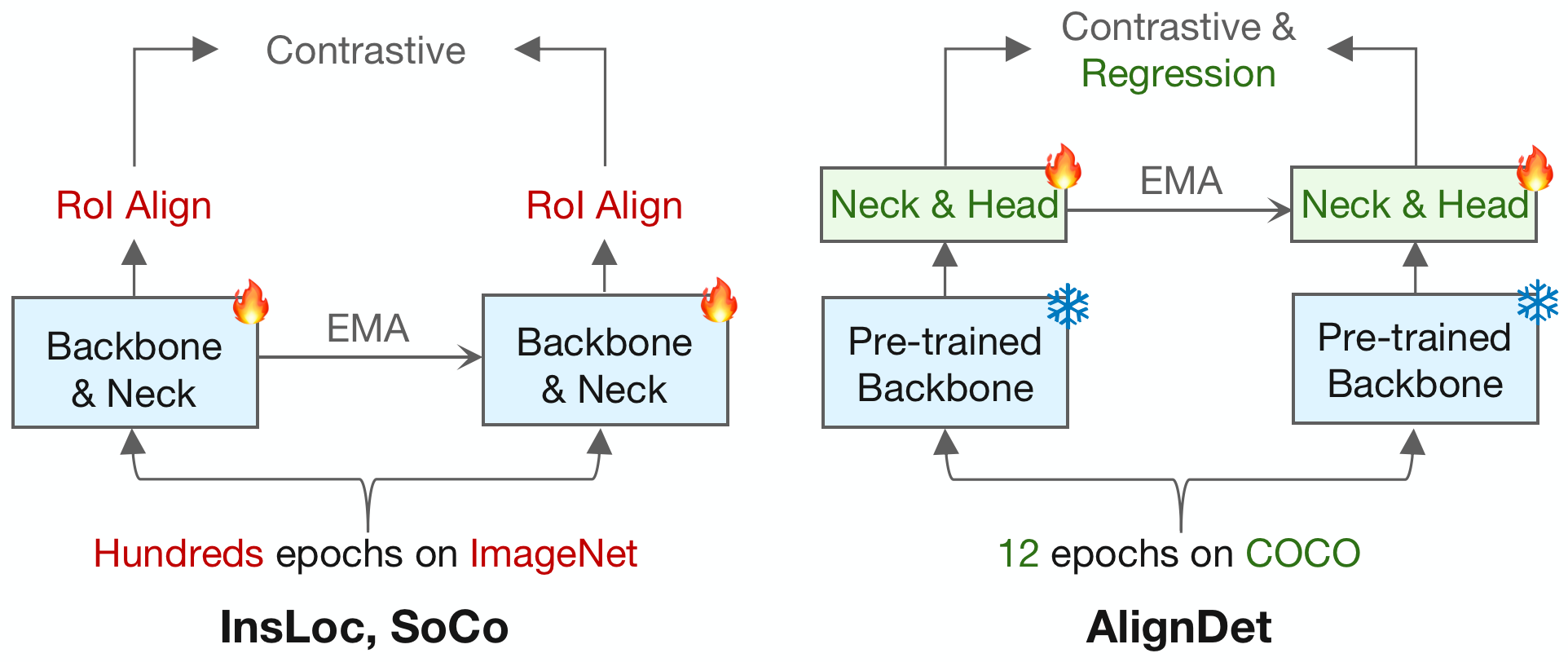}
    \caption{
    Compared with other box-level pre-training methods (\eg, InsLoc~\cite{insloc} and SoCo~\cite{soco}), our advantages are: 
    1) Data: AlignDet works well on COCO with only 12 epochs pre-training;
    2) Model: All the modules can be efficiently and fully pre-trained.
    3) Task: Both classification and regression knowledge are learned.
    }
    \label{fig:comparasion}
\end{figure}

As summarized in Table \ref{tab:comparison_discrepancy}, a series of works are proposed to bridge the gap between pre-training and fine-tuning processes in object detection.
The initial exploration is to construct dense-level contrastive learning to capture task-sensitive context for dense predictions \cite{densecl, self_emd, pixpro,insloc,soco,slotcon}.
Some researchers attempt to pre-train other detection modules, such as FPN~\cite{fpn,pixpro} and classification head~\cite{soco}.
However, 
these approaches require hundreds of epochs of pre-training on object-centric datasets and perform poorly when pre-training on COCO. In addition,
they neither pre-train all modules, such as the regression head in RetinaNet~\cite{retinanet} nor construct appropriate regression pre-training tasks, thus failing to resolve the model and task discrepancies, as illustrated in Figure~\ref{fig:comparasion}.
UP-DETR~\cite{up_detr} and DETReg~\cite{detreg} pre-train the entire DETR-like detectors by introducing DETR-sensitive pretext tasks. However, these tasks cannot be applied to other detectors since they rely heavily on the transformer-based decoder.
Although the above approaches can alleviate the gap to varying degrees, they cannot comprehensively solve the three discrepancies simultaneously or be generalized to various detectors. 
This leads us to ask: \textit{how to design a pre-training framework that can address the discrepancies in data, model, and task, and is applicable to all detection algorithms?}

To answer the above question, we propose AlignDet, a universal and pre-training framework for object detection.
AlignDet decouples the pre-training process into two stages, image-domain pre-training and box-domain pre-training. 
The image-domain pre-training optimizes the detection backbone to capture holistic visual abstraction, and the box-domain counterpart learns object-level concepts to initialize the parts out of the backbone.
The detector is optimized via box-level contrastive learning and coordinate-related regression losses.
It contributes to fully adapting to various detectors, further boosting the performance in the following fine-tuning process, as illustrated in Figure~\ref{fig:motivation}. The contributions of this work are summarized in four aspects:

\begin{itemize}[leftmargin=*]
\item \underline{\textit{New Insight}}: We point out that existing detection algorithms are constrained by the data, model, and task discrepancies between pre-training and fine-tuning.

\item \underline{\textit{Novel Method}}: We propose AlignDet to address this issue, which constructs detection-oriented pre-training by learning classification and regression knowledge. It decouples the pre-training into the image domain for the backbone and the box domain for other modules.

\item \underline{\textit{Efficiency and Pioneering}}: The module-based decouple takes full advantage of the existing pre-trained backbones to efficiently pre-train other modules. By incorporating self-supervised pre-trained backbones, we make the first attempt to fully pre-train various detectors using a completely unsupervised paradigm.

\item \underline{\textit{High Effectiveness}}: The comprehensive experiments demonstrate that AlignDet achieves significant performance improvements under various settings, including different detectors, backbones, data settings, and fine-tuning schedules.
\end{itemize}


\section{Related Work}
\label{sec:intro}

\paragraph{Object Detection.}
Current object detection algorithms can be divided into anchor-based, point-based, and query-based methods according to different prediction pipelines.
The anchor-based approaches generate multiple anchor boxes with pre-defined sizes and scales on each pixel of the feature maps~\cite{rcnn,fast_rcnn,faster_rcnn,mask_rcnn,ssd,retinanet}. Usually, they divide the training samples into positives and negatives by Intersection over Union (IoU).
The point-based methods aim to find the reference points that correspond to each object, which can be the center points of each instance~\cite{centernet, fcos}, pre-defined or self-learned key points~\cite{cornernet, extremenet, reppoints}.
Rather than leveraging the pre-defined priors in anchor-based and point-based methods, the query-based methods represent different objects~\cite{detr,deformable_detr,dino} through a set of learnable queries.

\paragraph{Self-supervised Pre-training.}
Self-supervised learning fully utilizes massive unlabeled data to learn structural data characteristics, and the pre-trained weights are transferred into downstream tasks to ensure good initialization~\cite{qin2022multi,wu2020fine,qin2023freeseg,wu2020tree,ppt,aqtc,gfm,inadequately}.
Numerous pretext tasks have been proposed for unsupervised pre-training, such as feature clustering~\cite{deep_cluster}, colorization~\cite{pretask_colorization}, context prediction~\cite{pretask_context}, rotation prediction~\cite{pretask_rotation} and image inpainting~\cite{pretask_inpainting}.
On the one hand, contrastive learning captures good representation by maximizing the similarity of different views from the same instance~\cite{moco,mocov2,simclr,byol,simsiam},  which achieves competitive performance on multiple downstream tasks.
On the other hand, Mask Image Modeling (MIM) has recently attracted increasing attention in self-supervised learning. MIM does not require specific data augmentation and more robust for downstream tasks~\cite{beit,ibot,simmim,mae}.

\paragraph{Self-supervised Pre-training for Detection.}
Although unsupervised pre-training has shown competitive results on object detection, there exists a series of inconsistencies in directly transferring image-level pre-trained knowledge to dense-level downstream tasks.
To bridge the gap between pre-training and fine-tuning, some approaches propose dense-level contrastive learning to explore the local feature similarity between different views~\cite{densecl,pixpro,self_emd,densesiam,scrl}.
Some researchers reveal that merely pre-training the backbone is insufficient, and they attempt to pre-training other common modules, such as FPN~\cite{fpn, soco}.
However, these methods require expensive pre-training from scratch, and other key modules in detectors (such as the regression head) remain randomly initialized.
On the other hand, UP-DETR~\cite{up_detr} and DETReg~\cite{detreg} pre-train the entire DETR-like detectors by introducing the region matching and feature reconstruction pretext tasks.
Although these approaches can pre-train the whole model adequately, the DETR-oriented pretext tasks cannot be directly applied to other detection paradigms. In contrast, AlignDet achieves efficient and adequate self-supervised pre-training for various detectors.

\section{Methodology}
\label{sec:method}


Recent works extend the ‘pre-training and fine-tuning’ paradigm via constructing unsupervised pre-training pretext tasks, resulting in higher performance than supervised pre-trained counterparts~\cite{moco,pixpro}.
However, compared to the detection process, the current pre-training paradigm has inconsistencies in data, model, and task, as illustrated in Figure~\ref{fig:motivation}. Although these inconsistencies can be alleviated by training on large-scale labeled datasets~\cite{obj365,bigdet,openimages}, it requires enormous computing resources and manual annotation costs. These problems and limitations inspire us to propose AlignDet, a universal and self-supervised framework for bridging the discrepancies between pre-training and fine-tuning in object detection.

\begin{figure*}[t!]\centering
    \includegraphics[width=1.0\linewidth]{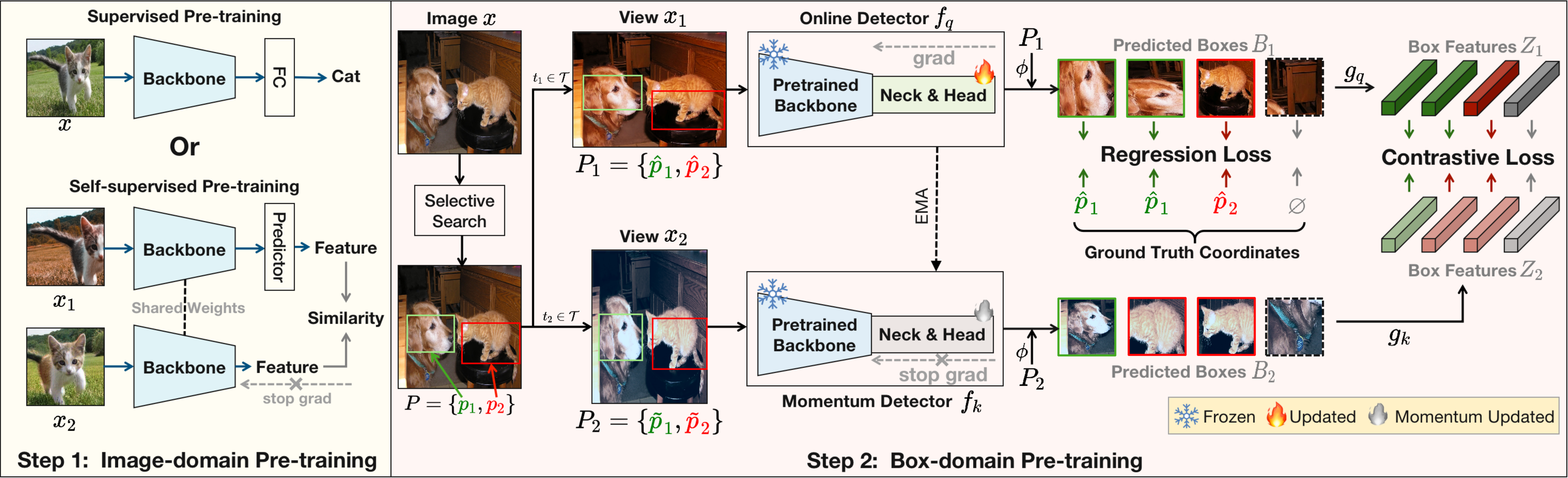}
    \caption{
    The pre-training pipeline of AlignDet.
Both supervised and self-supervised pre-training can be employed in the image-domain stage to capture holistic visual concepts.
For the box domain pre-training, selective search is first adopted to generate unsupervised proposals as pseudo labels, then each proposal is augmented to construct two views with different scales and transformations.  Each predicted box is used to construct contrastive learning and coordinated-related losses for adapting to detection-oriented tasks.
    }
    \label{fig:pipeline}
\end{figure*}

The whole pre-training pipeline is summarized in Figure~\ref{fig:pipeline}.
In the following subsections, we introduce the image-domain pre-training and box-domain pre-training in Section~\ref{subsec:image_pretrain} and Section~\ref{subsec:box_pretrain}, respectively. 
We provide pseudocode in Algorithm~\ref{code:aligndet} for a more intuitive understanding of the AlignDet pipeline, and the comparison with other methods on technical details in the \textbf{supplementary material}.

\subsection{Image-domain Pre-training}
\label{subsec:image_pretrain}
Image-domain pre-training optimizes the backbone to extract high-level semantic abstraction for the subsequent box-domain pre-training, as shown in Step 1 on the left of Figure~\ref{fig:pipeline}.
On the one hand, given an image $x$,  the backbone can be pre-trained with a classifier and classification category in the fully-supervised setting.
On the other hand, recently arisen unsupervised learning algorithms~\cite{moco,mocov2,simsiam} help to capture more generalized representations with the aid of massive unlabeled data.
Take SimSiam~\cite{simsiam} as an example, two views $x_1$ and $x_2$ are constructed from the input image with different data augmentation.
The backbone can learn generalized representations by maximizing the similarity of different views from the same instance.
And the predictor and stop gradient are leveraged to prevent the mode collapse~\cite{simsiam}.

Image-domain pre-training is usually conducted on large-scale image classification datasets such as ImageNet~\cite{imagenet}, in which each sample mainly contains one or a few salient objects in the center of the image. 
\textit{It exists an apparent gap because the pre-training procedure has no access to the target domain data, which often contains multiple objects with different scales and locations. 
Furthermore, the detection head is still randomly initialized and the regression task is not explicitly learned in this image-domain pre-training.
To this end, we design the box-domain pre-training to bridge these discrepancies.}

\subsection{Box-domain Pre-training}
\label{subsec:box_pretrain}

\paragraph{Self-supervised Object Detection.}
As represented in Figure~\ref{fig:pipeline},  given an input image $x$, we first generate the unsupervised proposals $P = \{p_1, p_2, ..., p_n \} $ via selective search~\cite{selective_search}. 
Each proposal $ p_i \in P $ can be presented as a bounding box with coordinates $(\hat{\mathrm{x}}, \hat{\mathrm{y}}, \hat{\mathrm{w}}, \hat{\mathrm{h}})$, where $(\hat{\mathrm{x}}, \hat{\mathrm{y}})$ denotes the coordinates of the box center and $(\hat{\mathrm{w}}, \hat{\mathrm{h}})$ denotes the width and height.
Then the image $x$ is transformed into augmented views $x_1$ and $x_2$ by applying a transformation $t$ sampled from the set of image transformations $\mathcal{T}$. The coordinates of unsupervised proposals $P$ also change into $P_1$ and $P_2$ according to the corresponding image transformations.
In every iteration of the pre-training procedure, the online detector $f_q$ and momentum detector $f_k$ will generate their predicted boxes of the image at different views.
By regarding the unsupervised proposals $P_1, P_2$ as ground truth, we can obtain the predicted boxes $B_1, B_2$ by:
\begin{equation}
\resizebox{.9\hsize}{!}{$
B_1 = \phi(f_q^{reg}(x_1), P_1),\quad B_2 = \phi(f_k^{reg}(x_2), P_2) \label{Equ.1} 
$}
\end{equation}
where $f^{reg}(x)$ denotes the box coordinates predicted by the regression-related modules $f^{reg}$ based on image $x$. $\phi$ represents the target assignment operation, which conducts sample matching between $f_q^{reg}(x_1)$ and unsupervised proposals $P_1$, such as calculating IoU in anchor-based detectors\footnote{Some anchor-free detectors~\cite{fcos} use pre-defined points to accomplish the procedure and some query-based detectors~\cite {detr, deformable_detr} divide positives and negatives by minimizing regression and classification cost.}. 
Note that the label-matching method varies with different detectors and we keep the default assignment mechanism of the detector without any changes, ensuring our method can be easily applied to different types of detectors.
Different from assigning the category of the matched object in the dataset and calculating classification loss in formal supervised training, we assign $l\in \{ \varnothing, 1, ..., n \}$, the index of each proposal in $P$ to the paired output of online detector $f_q$ and momentum detector $f_k$ by $\phi$. Here $\varnothing$ stands for the background, which means that the output box did not match any proposal in $P$. Then each output box in $B_1$ and $B_2$ can be rewritten as $b=(\mathrm{x}, \mathrm{y}, \mathrm{w}, \mathrm{h},l)$.

\paragraph{Box-domain Contrastive Learning.}
After obtaining the coordinates of each predicted box, we can further obtain the corresponding features in different views by replacing the original supervised classification head in the detection head with the unsupervised head $f^{con}$:
\begin{equation}
\resizebox{.9\hsize}{!}{$
Z_1 = g_q(f^{con}_{q}(x_1, B_1)), \quad Z_2 = g_k(f^{con}_{k}(x_2, B_2))
$}
\end{equation}
where $f^{con}(x, B)$ denotes the extracted features of predicted boxes $B$ for contrastive learning in the box-domain pre-training procedure.
And $g$ is the feature projection module, here we follow the architecture in MoCo v2~\cite{mocov2}, \emph{i.e.}, a 2-layer MLP head with ReLU~\cite{relu}. 
Please note that $f^{con}$ and $f^{reg}$ are usually two different modules in the detection head, and here we do not draw in Figure~\ref{fig:pipeline} for brevity.

In this unsupervised pre-training procedure, contrastive learning is formulated by the principle that the box representation corresponding to the \textit{same} proposal should be similar and vice versa. Specifically, we define the set of query boxes as $Q = \{b \in B_1: l \neq \varnothing \}$. For each query box $q \in Q$, assuming its assigned proposal index is $i$ and the feature is $z_q$. The set of positive keys $K_{+}$ and negative keys $K_{-}$ for query feature $z_q$ can be constructed as:
\begin{equation}
\resizebox{.9\hsize}{!}{$
Z_{+} = \left\{z \in Z_1: l = i\right\},\quad Z_{-} = \left\{z \in Z_1 \cup Z_2: l \neq i\right\}
$}
\end{equation}

Then the box-domain contrastive loss $\mathcal{L}_{con}$ for all query boxes in $Q$ is defined as:
\begin{equation}
\resizebox{0.9\linewidth}{!}{%
$
    \mathcal{L}_{c o n}=-\sum_{q} \sum_{z_{+}} \log \frac{\exp \left(z_q \cdot z_{+} / \tau\right)}{\exp \left(z_q \cdot z_{+} / \tau\right)+\sum_{z_{-}} \exp \left(z_q \cdot z_{-} / \tau\right)}
$
}
\end{equation}
where $\tau$ is a temperature hyper-parameter that controls the difficulty of the task of contrastive learning~\cite{tau}, we set 0.5 in our paper for all methods.

\paragraph{Overall Loss.}
For a prediction box $q_i = (\mathrm{x}, \mathrm{y}, \mathrm{w}, \mathrm{h}, \mathrm{l})$, where $q_i \in Q$. 
$l$ is the index of its corresponding proposal in $P_1$ which  calculated by the label assign function $\phi$. And we define it as $\hat{p}_l = (\hat{\mathrm{x}}, \hat{\mathrm{y}}, \hat{\mathrm{w}}, \hat{\mathrm{h}})$,
The coordinate-related regression losses are calculated according to each detector, and we do not make any modifications during pre-training:
\begin{equation}
    \mathcal{L}_{reg} = \sum_{q} \sum_{k=1}^{K} \lambda_{k} \cdot \mathcal{L}_{k} ((\mathrm{x}, \mathrm{y}, \mathrm{w}, \mathrm{h}), (\hat{\mathrm{x}}, \hat{\mathrm{y}}, \hat{\mathrm{w}}, \hat{\mathrm{h}}))
\end{equation}
where $K$ is the number of coordinate-related losses, and $\lambda_{k}$ denotes the loss factor. $\mathcal{L}_{k}$ could be any regression loss like IoU loss in FCOS~\cite{fcos} or L1 loss in Mask R-CNN~\cite{mask_rcnn}.
Then the overall loss is the combination of box-level contrastive loss $\mathcal{L}_{con}$ and the coordinate-related losses $\mathcal{L}_{reg}$:
\begin{equation}
    \mathcal{L} = \lambda_{con} \cdot \mathcal{L}_{con} + \lambda_{reg} \cdot \mathcal{L}_{reg}
\end{equation}
where $\lambda_{con}, \lambda_{reg}$ are the loss hyper-parameters, and we keep the same with the default setting from the corresponding detector.
We freeze the backbone in the box-domain stage to avoid detectors over-fit to noisy pseudo labels.
Box-domain pre-training addresses data and task discrepancies by employing multi-object data to construct detection-oriented pretext tasks. By cooperating with image-domain pre-training, AlignDet contributes to pre-train all modules within the detector, thus solving the model discrepancy.

\section{Experiments}
\label{sec:exp}
\subsection{Settings}

\paragraph{Datasets.}
In the image-domain pre-training stage, both the ImageNet~\cite{imagenet} and COCO~\cite{coco} dataset can be used to optimize the backbone. And the box domain pre-training employs the box domain datasets that contain non-object-centric and multi-object images.
Unless otherwise specified, we pre-train all methods on the COCO  train2017 dataset~\cite{coco} without any labels, then evaluate the detection model on the COCO val2017 dataset.
We also follow ~\cite{mocov2} to fine-tune the detectors on the VOC 07+12 train-val set and evaluate on the VOC 07 test set.

\paragraph{Data Augmentation.}
For the box-domain pre-training data augmentation, we follow  SoftTeacher~\cite{soft_teacher} but remove the RandomCrop and other box-jitter transforms to ensure all objects exist in both views.
The image resolution is [640, 800] to construct the multi-scale proposals and objects in the pre-training procedure.
In the fine-tuning phase, we follow the default augmentation settings of different methods in mmdetection~\cite{mmdet}.
And we follow ~\cite{moco,byol,soco,pixpro,densecl,swav} to use multi-scale training for self-supervised backbones.

\paragraph{Pre-training and Fine-tuning Details.}
In the pre-training phase, we fix all parameters of the backbone and update the neck and head.
Other hyper-parameters and settings are set as default in mmdetection~\cite{mmdet} except the hyper-parameters of prediction sampling.
In the fine-tuning phase, all the pre-trained weights except for the projection module are used to initialize the object detection model.
Synchronized batch normalization~\cite{bn} is used in both backbone and FPN following previous work~\cite{moco,byol,soco,rethinking}, which helps calibrate magnitudes for pre-trained models~\cite{moco}.
All pre-training follows the default 1x (90k) schedule except 50 epochs for DETR.
We adjust the learning rate and weight decay following previous work~\cite{soco}.
If not specified, the supervised pre-trained ResNet-50~\cite{resnet} in PyTorch~\cite{pytorch} is used by default for both the pre-training and fine-tuning stages.
The hyper-parameter details of different methods and experiments are summarized in the supplementary material.


\begin{table*}[]
\begin{center}
\resizebox{\textwidth}{!}{%
\begin{tabular}{c|c|ccccc|ccccc}
\multirow{2}{*}{\textbf{Detector}} &
  \multirow{2}{*}{\textbf{Align}} &
  \multicolumn{5}{c|}{\textbf{COCO 12k}} &
  \multicolumn{5}{c}{\textbf{COCO 90k}}\\ \cline{3-12} 
 &
   &
  \multicolumn{1}{c|}{\textbf{1\% Data}} &
  \multicolumn{1}{c|}{\textbf{5\% Data}} &
  \multicolumn{1}{c|}{\textbf{10\% Data}} &
  \multicolumn{1}{c|}{\textbf{20\% Data}} &
  \textbf{100\% Data} &
  \multicolumn{1}{c|}{\textbf{1\% Data}} &
  \multicolumn{1}{c|}{\textbf{5\% Data}} &
  \multicolumn{1}{c|}{\textbf{10\% Data}} &
  \multicolumn{1}{c|}{\textbf{20\% Data}} &
  \textbf{100\% Data} \\ \hline
FCOS &
  \textbf{\xmark} &
  \multicolumn{1}{c|}{8.1} &
  \multicolumn{1}{c|}{16.5} &
  \multicolumn{1}{c|}{21.5} &
  \multicolumn{1}{c|}{23.6} &
  22.1 &
  \multicolumn{1}{c|}{7.3} &
  \multicolumn{1}{c|}{16.0} &
  \multicolumn{1}{c|}{20.0} &
  \multicolumn{1}{c|}{23.8} &
  36.6 \\
\rowcolor{gray!10}FCOS &
  \textbf{\cmark} &
  \multicolumn{1}{c|}{\textbf{9.5 \textcolor[RGB]{96,177,87}{(+1.4)}}} &
  \multicolumn{1}{c|}{\textbf{18.6 \textcolor[RGB]{96,177,87}{(+2.1)}}} &
  \multicolumn{1}{c|}{\textbf{23.3 \textcolor[RGB]{96,177,87}{(+1.8)}}} &
  \multicolumn{1}{c|}{\textbf{26.3 \textcolor[RGB]{96,177,87}{(+2.7)}}} &
  \textbf{27.4 \textcolor[RGB]{96,177,87}{(+5.3)}} &
  \multicolumn{1}{c|}{\textbf{8.5 \textcolor[RGB]{96,177,87}{(+1.2)}}} &
  \multicolumn{1}{c|}{\textbf{17.3 \textcolor[RGB]{96,177,87}{(+1.3)}}} &
  \multicolumn{1}{c|}{\textbf{21.1 \textcolor[RGB]{96,177,87}{(+1.1)}}} &
  \multicolumn{1}{c|}{\textbf{25.1 \textcolor[RGB]{96,177,87}{(+1.3)}}} &
  \textbf{37.5 \textcolor[RGB]{96,177,87}{(+0.9)}}
\\ \hline
RetinaNet &
  \textbf{\xmark} &
  \multicolumn{1}{c|}{8.0} &
  \multicolumn{1}{c|}{17.8} &
  \multicolumn{1}{c|}{21.0} &
  \multicolumn{1}{c|}{23.0} &
  24.2 &
  \multicolumn{1}{c|}{8.3} &
  \multicolumn{1}{c|}{18.0} &
  \multicolumn{1}{c|}{22.2} &
  \multicolumn{1}{c|}{25.9} &
  36.3 \\
\rowcolor{gray!10}RetinaNet &
  \textbf{\cmark} &
  \multicolumn{1}{c|}{\textbf{9.8  \textcolor[RGB]{96,177,87}{(+1.8)}}} &
  \multicolumn{1}{c|}{\textbf{19.3 \textcolor[RGB]{96,177,87}{(+1.5)}}} &
  \multicolumn{1}{c|}{\textbf{23.5 \textcolor[RGB]{96,177,87}{(+2.5)}}} &
  \multicolumn{1}{c|}{\textbf{25.9 \textcolor[RGB]{96,177,87}{(+2.9)}}} &
  \textbf{26.3 \textcolor[RGB]{96,177,87}{(+2.1)}} &
  \multicolumn{1}{c|}{\textbf{9.9  \textcolor[RGB]{96,177,87}{(+1.6)}}} &
  \multicolumn{1}{c|}{\textbf{19.0 \textcolor[RGB]{96,177,87}{(+1.0)}}} &
  \multicolumn{1}{c|}{\textbf{23.1 \textcolor[RGB]{96,177,87}{(+0.9)}}} &
  \multicolumn{1}{c|}{\textbf{26.6 \textcolor[RGB]{96,177,87}{(+0.7)}}} &
  \textbf{37.3 \textcolor[RGB]{96,177,87}{(+1.0)}}
\\ \hline
Faster R-CNN &
  \textbf{\xmark} &
  \multicolumn{1}{c|}{9.2} &
  \multicolumn{1}{c|}{18.7} &
  \multicolumn{1}{c|}{24.2} &
  \multicolumn{1}{c|}{26.6} &
  27.3 &
  \multicolumn{1}{c|}{6.5} &
  \multicolumn{1}{c|}{14.0} &
  \multicolumn{1}{c|}{18.8} &
  \multicolumn{1}{c|}{24.1} &
  37.9 \\
\rowcolor{gray!10}Faster R-CNN &
  \textbf{\cmark} &
  \multicolumn{1}{c|}{\textbf{11.7  \textcolor[RGB]{96,177,87}{(+2.5)}}} &
  \multicolumn{1}{c|}{\textbf{21.2 \textcolor[RGB]{96,177,87}{(+2.5)}}} &
  \multicolumn{1}{c|}{\textbf{26.9 \textcolor[RGB]{96,177,87}{(+2.7)}}} &
  \multicolumn{1}{c|}{\textbf{29.6 \textcolor[RGB]{96,177,87}{(+3.0)}}} &
  \textbf{30.6 \textcolor[RGB]{96,177,87}{(+3.3)}} &
  \multicolumn{1}{c|}{\textbf{8.9 \textcolor[RGB]{96,177,87}{(+2.4)}}} &
  \multicolumn{1}{c|}{\textbf{16.2 \textcolor[RGB]{96,177,87}{(+2.2)}}} &
  \multicolumn{1}{c|}{\textbf{20.1 \textcolor[RGB]{96,177,87}{(+1.3)}}} &
  \multicolumn{1}{c|}{\textbf{25.2 \textcolor[RGB]{96,177,87}{(+1.1)}}} &
  \textbf{39.0 \textcolor[RGB]{96,177,87}{(+1.1)}}
\\ \hline
Mask R-CNN &
  \textbf{\xmark} &
  \multicolumn{1}{c|}{8.8} &
  \multicolumn{1}{c|}{19.1} &
  \multicolumn{1}{c|}{24.2} &
  \multicolumn{1}{c|}{26.5} &
  27.2 &
  \multicolumn{1}{c|}{7.6} &
  \multicolumn{1}{c|}{15.2} &
  \multicolumn{1}{c|}{20.0} &
  \multicolumn{1}{c|}{25.2} &
  38.3 \\
\rowcolor{gray!10}Mask R-CNN &
  \textbf{\cmark} &
  \multicolumn{1}{c|}{\textbf{12.4 \textcolor[RGB]{96,177,87}{(+3.6)}}} &
  \multicolumn{1}{c|}{\textbf{22.4 \textcolor[RGB]{96,177,87}{(+3.3)}}} &
  \multicolumn{1}{c|}{\textbf{27.4 \textcolor[RGB]{96,177,87}{(+3.2)}}} &
  \multicolumn{1}{c|}{\textbf{30.1 \textcolor[RGB]{96,177,87}{(+3.6)}}} &
  \textbf{30.5 \textcolor[RGB]{96,177,87}{(+3.3)}} &
  \multicolumn{1}{c|}{\textbf{9.5 \textcolor[RGB]{96,177,87}{(+1.9)}}} &
  \multicolumn{1}{c|}{\textbf{16.6 \textcolor[RGB]{96,177,87}{(+1.4)}}} &
  \multicolumn{1}{c|}{\textbf{20.8 \textcolor[RGB]{96,177,87}{(+0.8)}}} &
  \multicolumn{1}{c|}{\textbf{25.9 \textcolor[RGB]{96,177,87}{(+0.7)}}} &
  \textbf{39.4 \textcolor[RGB]{96,177,87}{(+1.1}}
\\ \hline
\multirow{1}{*}{\textbf{Detector}} &
\multirow{1}{*}{\textbf{Align}} &
  \multicolumn{5}{c|}{\textbf{COCO 50 epochs}} &
  \multicolumn{5}{c}{\textbf{COCO 100 epochs}} \\ \hline
DETR &
  \textbf{\xmark} &
  \multicolumn{1}{c|}{7.6} &
  \multicolumn{1}{c|}{18.9} &
  \multicolumn{1}{c|}{24.1} &
  \multicolumn{1}{c|}{29.7} &
  35.0 &
  \multicolumn{1}{c|}{7.7} &
  \multicolumn{1}{c|}{19.5} &
  \multicolumn{1}{c|}{25.0} &
  \multicolumn{1}{c|}{30.4} &
  38.4 \\
\rowcolor{gray!10}DETR &
  \textbf{\cmark} &
  \multicolumn{1}{c|}{\textbf{11.2 \textcolor[RGB]{96,177,87}{(+3.6)}}} &
  \multicolumn{1}{c|}{\textbf{21.4 \textcolor[RGB]{96,177,87}{(+2.5)}}} &
  \multicolumn{1}{c|}{\textbf{26.1 \textcolor[RGB]{96,177,87}{(+2.0)}}} &
  \multicolumn{1}{c|}{\textbf{30.9 \textcolor[RGB]{96,177,87}{(+1.2)}}} &
  \textbf{37.3 \textcolor[RGB]{96,177,87}{(+2.3)}} &
  \multicolumn{1}{c|}{\textbf{10.7 \textcolor[RGB]{96,177,87}{(+3.0)}}} &
  \multicolumn{1}{c|}{\textbf{21.2 \textcolor[RGB]{96,177,87}{(+1.7)}}} &
  \multicolumn{1}{c|}{\textbf{26.0 \textcolor[RGB]{96,177,87}{(+1.0)}}} &
  \multicolumn{1}{c|}{\textbf{31.2 \textcolor[RGB]{96,177,87}{(+0.8)}}} &
  \textbf{39.0 \textcolor[RGB]{96,177,87}{(+0.6)}}
\end{tabular}}
\end{center}
\vspace{-0.1cm}
\caption{
With only 12 epochs pre-training on COCO for modules out of backbone, AlignDet achieves consistent improvements across different detectors, training strategies, and data sizes. All the results are conducted with a supervised pre-trained ResNet-50 backbone.}
\vspace{-0.1cm}
\label{tab:coco_12k_1x}
\end{table*}

\begin{algorithm}[t]
\caption{AlignDet Pseudocode, PyTorch-like}
\vspace{-0.2cm}
\label{code:aligndet}
\definecolor{codeblue}{rgb}{0.25,0.5,0.5}
\definecolor{codekw}{rgb}{0.85, 0.18, 0.50}
\lstset{
  backgroundcolor=\color{white},
  basicstyle=\fontsize{6.5pt}{6.5pt}\ttfamily\selectfont,
  columns=fullflexible,
  breaklines=true,
  captionpos=b,
  commentstyle=\fontsize{6.5pt}{6.5pt}\color{codeblue},
  keywordstyle=\fontsize{6.5pt}{6.5pt}\color{codekw},
}
\begin{lstlisting}[language=python]
# x: input images
# p: selective search proposals
# aug: independent random augmentation

for x, p in data_loader:
    (x1, p1), (x2, p2) = aug(x, p), aug(x, p) # augmentation
    x1, x2 = backbone(x1), backbone(x2)  # frozen backbone
    x1, x2 = neck_q(x1), neck_k(x2)  # feature pyramid

    # proposals as pseudo labels, boxes are predicted
    b1, b2 = head_q.f_reg(x1, p1), head_k.f_reg(x2, p2)
    z1, z2 = head_q.f_con(x1, b1), head_k.f_con(x2, b2)
    z1, z2 = g_q(z1), g_k(z2)  # feature projection

    L = loss_con(z1, z2) + loss_reg(b1, b2, p1, p2)  # losses
    ema_update(neck_q, neck_k, head_q, head_k)
\end{lstlisting}
\vspace{-0.2cm}
\end{algorithm}

\subsection{Experimental Results}
\paragraph{Detectors and Data Settings.}
To demonstrate the generalization of our approach, we leverage AlignDet to pre-train different detectors, including FCOS~\cite{fcos} (single-stage and point-based), RetinaNet~\cite{retinanet} (single-stage and anchor-based),  Faster R-CNN and Mask R-CNN~\cite{mask_rcnn} (two-stage and anchor-based), and DETR~\cite{detr} (query-based). 
Here Faster R-CNN uses RoI Align following previous work~\cite{mmdet,moco}, and Mask R-CNN is fine-tuned with both detection and instance segmentation annotations following~\cite{soco}.
We random sample 1\%, 5\%, 10\%, and 20\% images from COCO train2017 set as the fine-tuning data. We provide 5 different data folds for each low-data setting, and the final performance is the average of all results.
To avoid over-fitting and demonstrate the advantage of faster convergence, besides the standard 1x (90k iterations) training schedule, we also report the results of 12k iterations in a low-data regime following~\cite{detco}.
Table~\ref{tab:coco_12k_1x} summarizes the detection performance of various detectors with different data settings.

\begin{table}[]
\begin{center}
\resizebox{\columnwidth}{!}{%
\begin{tabular}{c|c|c|c|c}
\textbf{Backbone} & \textbf{Align} & \textbf{Schedule} & \textbf{AP$^{bb}$} & \textbf{AP$^{mk}$}  \\
\hline
MobileNet v2  & \xmark & 1x & 30.1  & 27.2  \\
\rowcolor{gray!10} MobileNet v2  & \cmark & 1x & \textbf{31.3 \textcolor[RGB]{96,177,87}{(+1.2)}}  & \textbf{27.9 \textcolor[RGB]{96,177,87}{(+0.7)}}  \\ \hline
ResNet-18        & \xmark & 1x & 34.5  & 31.1  \\
\rowcolor{gray!10}ResNet-18        & \cmark & 1x & \textbf{35.7 \textcolor[RGB]{96,177,87}{(+1.2)}}  & \textbf{31.9 \textcolor[RGB]{96,177,87}{(+0.8)}}  \\
\hline
ResNet-50        & \xmark & 1x & 38.3                 & 34.3                 \\
\rowcolor{gray!10}ResNet-50        & \cmark & 1x & \textbf{39.4 \textcolor[RGB]{96,177,87}{(+1.1)}} & \textbf{35.3 \textcolor[RGB]{96,177,87}{(+1.0)}} \\
\hline
Swin-Small & \xmark & 1x & 46.6                 & 41.5                 \\
\rowcolor{gray!10}Swin-Small & \cmark & 1x & \textbf{47.5 \textcolor[RGB]{96,177,87}{(+0.9)}}  & \textbf{41.8 \textcolor[RGB]{96,177,87}{(+0.3)}} \\
\hline
Swin-Small & \xmark & 3x & 48.2                 & 43.2                 \\
\rowcolor{gray!10}Swin-Small & \cmark & 3x & \textbf{49.1 \textcolor[RGB]{96,177,87}{(+0.9)}} & \textbf{43.4 \textcolor[RGB]{96,177,87}{(+0.2)}} \\
\hline
Swin-Base  & \xmark & 1x & 48.8                 & 43.1                 \\
\rowcolor{gray!10}Swin-Base  & \cmark & 1x & \textbf{49.6 \textcolor[RGB]{96,177,87}{(+0.8)}} & \textbf{43.6 \textcolor[RGB]{96,177,87}{(+0.5)}} \\
\hline
\textbf{Method} & \textbf{Align} & \textbf{Schedule} & \textbf{AP$^{bb}$} & \textbf{AP$^{mk}$}  \\
\hline
SimMIM$^\dag$ (Swin-B) & \xmark & 3x & 51.0                 & 45.1                 \\
\rowcolor{gray!10}SimMIM (Swin-B) & \cmark & 3x & \textbf{51.6 \textcolor[RGB]{96,177,87}{(+0.6)}} & \textbf{45.8 \textcolor[RGB]{96,177,87}{(+0.7)}} \\
\hline
CBNet v2$^\dag$ (Swin-L) & \xmark & 1x & 57.3                 & 49.7                 \\
\rowcolor{gray!10}CBNet v2 (Swin-L) & \cmark & 1x & \textbf{57.8 \textcolor[RGB]{96,177,87}{(+0.5)}} & \textbf{50.1 \textcolor[RGB]{96,177,87}{(+0.4)}}

\end{tabular}%
}
\end{center}
\caption{
AlignDet achieves consistent improvements on different backbones and the SOTA detector CBNet v2. $\dag$ denotes our reproduced results with official code or models.
}
\label{tab:different backbones}
\end{table}

\begin{table}[]
\begin{center}
\resizebox{\columnwidth}{!}{%
\begin{tabular}{c|c|c|c|c}
\textbf{Detector} & \textbf{Align} & \textbf{AP} & \textbf{AP$_{50}$} & \textbf{AP$_{75}$} \\
\hline
FCOS         & \xmark    &  52.6  &  79.6   &  57.1    \\
\rowcolor{gray!10}FCOS         & \cmark    &  \textbf{53.4 \textcolor[RGB]{96,177,87}{(+0.8)}}  & \textbf{80.2 \textcolor[RGB]{96,177,87}{(+0.6)}}    &  \textbf{65.2 \textcolor[RGB]{96,177,87}{(+8.1)}}   
\\ \hline
RetinaNet    & \xmark    &  54.4  & 79.3    &  58.7    \\
\rowcolor{gray!10}RetinaNet    & \cmark    &  \textbf{56.0 \textcolor[RGB]{96,177,87}{(+1.6)}}  & \textbf{80.4 \textcolor[RGB]{96,177,87}{(+1.1)}}    & \textbf{61.0 \textcolor[RGB]{96,177,87}{(+2.3)}}  
\\ \hline
Faster R-CNN   & \xmark    & 53.5   & 81.4    &  58.2    \\
\rowcolor{gray!10}Faster R-CNN    & \cmark    &  \textbf{57.8 \textcolor[RGB]{96,177,87}{(+4.3)}}  & \textbf{82.9 \textcolor[RGB]{96,177,87}{(+1.6)}}    & \textbf{64.7 \textcolor[RGB]{96,177,87}{(+6.5)}}  
\\ \hline
DETR         & \xmark    & 52.1   & 76.8    &  54.9    \\
\rowcolor{gray!10}DETR         & \cmark    &  \textbf{58.2 \textcolor[RGB]{96,177,87}{(+6.1)}}  & \textbf{81.1 \textcolor[RGB]{96,177,87}{(+4.3)}}    & \textbf{62.9 \textcolor[RGB]{96,177,87}{(+8.0)}} 
\end{tabular}%
}
\end{center}
\caption{
Transfer learning results on Pascal VOC benchmark. The prior knowledge learned by AlignDet pre-training can effectively help the downstream detection task.}
\label{tab:voc}
\end{table}

Compared with ImageNet pre-training, our AlignDet has significantly improved performance under different settings. Even with complete data (100\%) and fine-tuning schedule (90k), there are nearly 1.0 mAP improvements for different detection algorithms.
AlignDet achieves a noticeable performance improvement in shorter training iterations and data settings.
For example, AlignDet helps Mask R-CNN bring at least +3.2 mAP improvement in 12k iterations under diverse data protocols.
Furthermore, the detectors suffer from the over-fitting issue in the low-data regime, in which Mask R-CNN with 90k iterations (15.2 mAP) is 3.9 mAP lower than with 12k iterations (19.1 mAP) at 5\% data. 
However, AlignDet still can obtain 1.4 mAP improvement under this dilemma, which means that the knowledge introduced by AlignDet pre-training can help the model avoid over-fitting.
\textit{Note that these non-trivial improvements are achieved by highly efficient pre-training. Only modules other than the backbone will be updated with 12 epochs pre-training on a relatively small COCO dataset, which is more efficient than other methods, as shown in Figure~\ref{fig:comparasion}.}

\paragraph{Detection Backbones and SOTA model.}
As shown in Table~\ref{tab:different backbones}, we validate the effectiveness of AlignDet on various backbones with the Mask R-CNN framework, including MobileNet v2~\cite{mbv2}, ResNets~\cite{resnet}, and Swin Transformers~\cite{swin}.
The experimental results show that our AlignDet can improve various backbones effectively.
For example, AlignDet improves mAP by +1.2, +1.1, and +0.8 on supervised pre-trained MobileNet v2, ResNet-50, and Swin-Base, respectively. 
To further verify the effectiveness of AlignDet, we conducted advanced experiments on a mask image modeling pre-trained backbone (SimMIM~\cite{simmim} pre-trained Swin-Base), and the state-of-the-art detection model without additional detection data (CBNet v2~\cite{cbnet} with Swin-Large backbone). Both of them use strong data augmentation and train to convergence. The experiments demonstrate that AlignDet achieves consistent and impressive performance improvements on the SOTA model (CBNet v2) and advanced techniques (SimMIM).

\paragraph{Generalization Analysis.}
Table~\ref{tab:voc} shows the transfer results from COCO pre-training to PASCAL VOC fine-tuning.
The VOC dataset has fewer categories compared to the COCO dataset, resulting in easier classification by the model. Therefore, we should focus more on the metric with a higher IoU threshold, \emph{i.e.}, AP${_{75}}$.
The considerable improvement of AlignDet pre-training on AP$_{75}$ shows that our pre-training makes the predicted coordinates more accurate, which also reveals that the prior knowledge learned by AlignDet pre-training can be effectively transferred into downstream detection datasets and tasks.

\begin{table}[]
\begin{center}
\resizebox{\linewidth}{!}{%
\begin{tabular}{c|c|c|ccc|ccc}
\multirow{2}{*}{\textbf{Method}}
& \multirow{2}{*}{\textbf{Dataset}}
& \multirow{2}{*}{\textbf{Epochs}}
& \multicolumn{3}{c|}{\textbf{Mask R-CNN}}
& \multicolumn{3}{c}{\textbf{RetinaNet}}      \\
\cline{4-9}
 &  &   
& \textbf{AP} & \textbf{AP$_{50}$} & \textbf{AP$_{75}$}
& \textbf{AP} & \textbf{AP$_{50}$} & \textbf{AP$_{75}$} \\
\hline
ReSim & ImageNet & 400  & 40.3 & 60.6 & 44.2 & -    & -    & -    \\
InsLoc  & ImageNet &  400    & 42.0 & 62.3 & 45.8 & -    & -    & -    \\
SoCo  & ImageNet &  400  & \textbf{43.0} & \textbf{63.3} & \textbf{47.1} & 38.3 & 57.2 & 41.2 \\
SoCo   & COCO & 530   & 40.6 & 61.1 & 44.4 & - & - & - \\
\Xhline{1pt}
Supervised      & ImageNet &  90   & 39.7 & 59.5 & 43.3 & 37.4 & 56.6 & 39.7 \\
DetCo   & ImageNet & 200      & 40.7 & 60.5 & 44.6 & 38.0 & 57.0 & 40.5 \\
DenseCL  & ImageNet & 200   & 40.3 & 59.9 & 44.3 & 37.5 & 56.0 & 39.8 \\
DenseCL  & COCO & 800   & 39.6 & 59.3 & 43.3 & - & - & - \\
Self-EMD  & ImageNet & 300  & 40.0 & 60.4 & 44.0 & - & - & - \\
Self-EMD  & COCO+ & 800  & 40.4 & 61.1 & 43.7 & 37.4 & 56.5 & 39.7 \\
SlotCon   & COCO  & 800  & 41.0 & 61.1 & 45.0 & - & - & - \\
MoCo v2  & ImageNet & 800    & 40.3 & 59.9 & 43.9 & 37.8 & 56.4 & 40.4 \\
\rowcolor{gray!10}+AlignDet  & COCO & 12   & 41.0 & 61.0 & 44.9 & 38.4 & 57.1 & 41.3 \\
\hline
PixPro{$^\dag$}   & ImageNet & 400    & 40.9 & 60.4 & 44.8 & 38.4 & 57.0 & 41.4 \\
\rowcolor{gray!10}+AlignDet  & COCO & 12  & 41.7 & 61.7 & 45.5 & \textbf{39.0} & \underline{57.5} & \textbf{41.9} \\
\hline
SwAV    & ImageNet & 800     & 41.6 & 62.2 & 45.8 & 37.1 & 56.8 & 39.5 \\
\rowcolor{gray!10}+AlignDet  & COCO & 12  & \underline{42.3} & \underline{62.5} & \underline{46.7} & \underline{38.5} & \textbf{58.2} & \underline{41.0}
\end{tabular}%
}
\end{center}
\caption{
AlignDet achieves competitive results compared to other methods, with only 12 epochs pre-training on COCO. $\dag$ denotes our reproduced results with officially released model and code.}
\label{tab:selfsup_backbone}
\end{table}

\begin{table}[]
\begin{center}
\resizebox{\columnwidth}{!}{%
\begin{tabular}{c|c|c|c|c|c|c}
\textbf{Method} & \textbf{Backbone} & \textbf{Dataset} & \textbf{Epochs} & \textbf{AP} & \textbf{AP$_{50}$} & \textbf{AP$_{75}$} \\ \hline
From Scratch & Sup. R50 & -        & -  & 39.5 & 60.3 & 41.4 \\
From Scratch & SwAV R50 & -        & -  & 39.7 & 60.3 & 41.7 \\ \hline
UP-DETR~\cite{up_detr}      & SwAV R50 & ImageNet & 60 & 40.5 & 60.8 & 42.6 \\
DETReg$^\dag$~\cite{detreg}       & SwAV R50 & COCO     & 50 & 39.8    & 61.0    & 41.6    \\ \hline
AlignDet         & Sup. R50 & COCO     & 50 & 41.0    & 61.9    & 43.1    \\
AlignDet         & SwAV R50 & COCO     & 50 & \textbf{41.4} & \textbf{62.1} & \textbf{43.8}
\end{tabular}%
}
\end{center}
\caption{
Comparasion of AlignDet with DETR-specific methods. AlignDet achieves SOTA performance with the same or fewer pre-training. All results are fine-tuned with 150 epochs on COCO.
}
\label{tab:detr}
\end{table}

\paragraph{Compare with Other Self-supervised Methods.}
As summarized in Table~\ref{tab:selfsup_backbone}, we also validate the effectiveness of AlignDet on the self-supervised ResNet-50 backbones with Mask R-CNN and RetinaNet. All the experiments follow the 1x fine-tuning strategy defined in~\cite{moco}.
AlignDet can significantly improve the detector's performance under various pre-trained backbones with only 12 epochs of box-domain pre-training on COCO.
For example, AlignDet boosts the performance by +0.6 mAP under the RetinaNet with PixPro pre-trained backbone.
AlignDet does not outperform SoCo on the Mask R-CNN because SoCo conducts end-to-end pre-training on most modules (except RPN) for a long time (400 epochs) on the full ImageNet (\textbf{5.3x larger than COCO}). When SoCo is pre-trained on COCO, its performance drops significantly, While AlignDet only needs 12 epochs (\textbf{33.3$\times$ acceleration}) on COCO to pre-train the modules out of the backbone.
It owes to AlignDet decoupling the pre-training process and taking advantage of the existing pre-trained backbones to accelerate convergence.
In addition, SoCo cannot adequately pre-train detectors decoupled from regression and classification branches (\emph{e.g.}, RetinaNet), leading to sub-optimal performance. 
On the contrary, AlignDet achieves significant performance improvement with various detectors and backbones. 
We also compare with other self-supervised methods explicitly designed for DETR, as shown in Table~\ref{tab:detr}.
AlignDet shows noticeable improvements over these DETR-specific methods. For example, AlignDet with 50 epochs pre-training on COCO surpasses UP-DETR with 60 epochs on ImageNet by +1.1 mAP.
\textit{Notably, by incorporating the self-supervised backbone, we can pre-train all modules within various detectors in a completely unsupervised paradigm.}

\begin{table}[!]
\begin{center}
\resizebox{\columnwidth}{!}{%
\begin{tabular}{c|ccc|ccc}
\textbf{\makecell{Fine-tuning\\Schedule}}  & \textbf{AP$^{bb}$} & \textbf{AP$^{bb}_{50}$} & \textbf{AP$^{bb}_{75}$} & \textbf{AP$^{mk}$} & \textbf{AP$^{mk}_{50}$} & \textbf{AP$^{mk}_{75}$}  \\ \hline
1x  & 38.3  & 58.0  & 42.1  & 34.3  & 54.9  & 36.6  \\
2x  & 38.8  & 58.4  & 42.4  & 34.7  & 55.4  & 37.2 \\
3x  & 39.0  & 58.7  & 42.9  & 35.0  & 55.6  & 37.7 \\
4x  & 39.2  & 59.5  & 42.9  & 35.6  & 56.5  & 38.1 \\
\hline
1x (Ours)  & \textbf{39.4 \textcolor[RGB]{96,177,87}{(+1.1)}} & \textbf{59.2 \textcolor[RGB]{96,177,87}{(+1.2)}} & \textbf{43.2 \textcolor[RGB]{96,177,87}{(+1.1)}} & \textbf{35.3 \textcolor[RGB]{96,177,87}{(+1.0)}} & \textbf{56.1 \textcolor[RGB]{96,177,87}{(+1.2)}} & \textbf{37.7 \textcolor[RGB]{96,177,87}{(+1.1)}}  \\
2x (Ours)  & \textbf{39.8 \textcolor[RGB]{96,177,87}{(+1.0)}} & \textbf{59.3 \textcolor[RGB]{96,177,87}{(+0.9)}} & \textbf{43.3 \textcolor[RGB]{96,177,87}{(+0.9)}} & \textbf{35.3 \textcolor[RGB]{96,177,87}{(+0.6)}} & \textbf{56.4 \textcolor[RGB]{96,177,87}{(+1.0)}} & \textbf{37.6 \textcolor[RGB]{96,177,87}{(+0.4)}}
\end{tabular}%
}
\end{center}
\caption{
Results with longer fine-tuning epochs. Training the baseline (ImageNet pre-trained backbone) with longer epochs until convergence (4x) cannot bring similar improvements as AlignDet.
}
\vspace{-0.1cm}
\label{tab:longer_schedule}
\end{table}

\paragraph{Training Schedules.}
We conduct experiments with diverse training schedules to eliminate the concern that the performance gain from pre-training is due to longer training time. 
As shown in Table~\ref{tab:longer_schedule}, the performance of AlignDet with 1x pre-training and fine-tuning is even 0.2 mAP higher than the 4$\times$ fine-tuning result of ImageNet pre-training, which indicates that the performance gain is mainly due to the design of AlignDet rather than the longer training time.

\subsection{Ablation Study}
We conduct a series of ablation studies to further understand the advantages of bridging the discrepancies between pre-training and fine-tuning in terms of data, model, and task.
All experiments are conducted on Faster R-CNN with ResNet-50~\cite{resnet} initialized by the supervised pre-trained weights.
The pre-training and fine-tuning procedure adopts the standard 1x training schedule on COCO~\cite{coco}.

\paragraph{Data Discrepancy: Do data characteristics and domains matter in pre-training?}
Yes. Table~\ref{tab:ablation_data} demonstrates the effectiveness of bridging the data discrepancy.
Leveraging a single object-centric dataset such as ImageNet to pre-train the detector backbone will result in a poor performance of 37.9 mAP.
By introducing box-domain pre-training in the COCO dataset, we can fully use multi-object datasets to mitigate inconsistencies in data characteristics, improving +1.1 mAP.
To further verify the influence of the data domain, we also employ selective search to construct the ImageNet Subset, where the number of objects is consistent with COCO. There is still a noticeable performance gap between pre-training on ImageNet Subset and COCO, which shows that the data domain is essential in pre-training.

\paragraph{Model Discrepancy: Are all detector modules necessary to conduct pre-training?}
To answer this question, we load part of the pre-trained weights and remain other modules randomly initialization to investigate the benefit of pre-training different modules.
As summarized in Table~\ref{tab:ablation_model}, we can obtain the following observations.
i) Additional pre-training Neck improves the mAP by 0.5, of which the most significant improvement comes from AP$_{75}$ and AP$_{l}$. It indicates that Neck pre-training brings more accurate prediction boxes and larger size objects.
ii)  An interesting finding is that compared with the variants that the RPN and Head are all randomly initialized, merely pre-training the RPN causes performance decreases by 0.2 mAP.
It may be because the randomly initialized head module cannot correctly handle the coarse results from the RPN.
iii)  On the contrary, if we pre-train other modules and maintain random initialization for RPN,  the final performance is only 0.2 mAP lower than that of full pre-training. It may be because the RPN module only occupies 1.39\%  parameters within the whole detector.
In conclusion, complete pre-training of the whole detector can bring the best downstream performance.

\begin{figure}[t]\centering
    \includegraphics[width=1.0\linewidth]{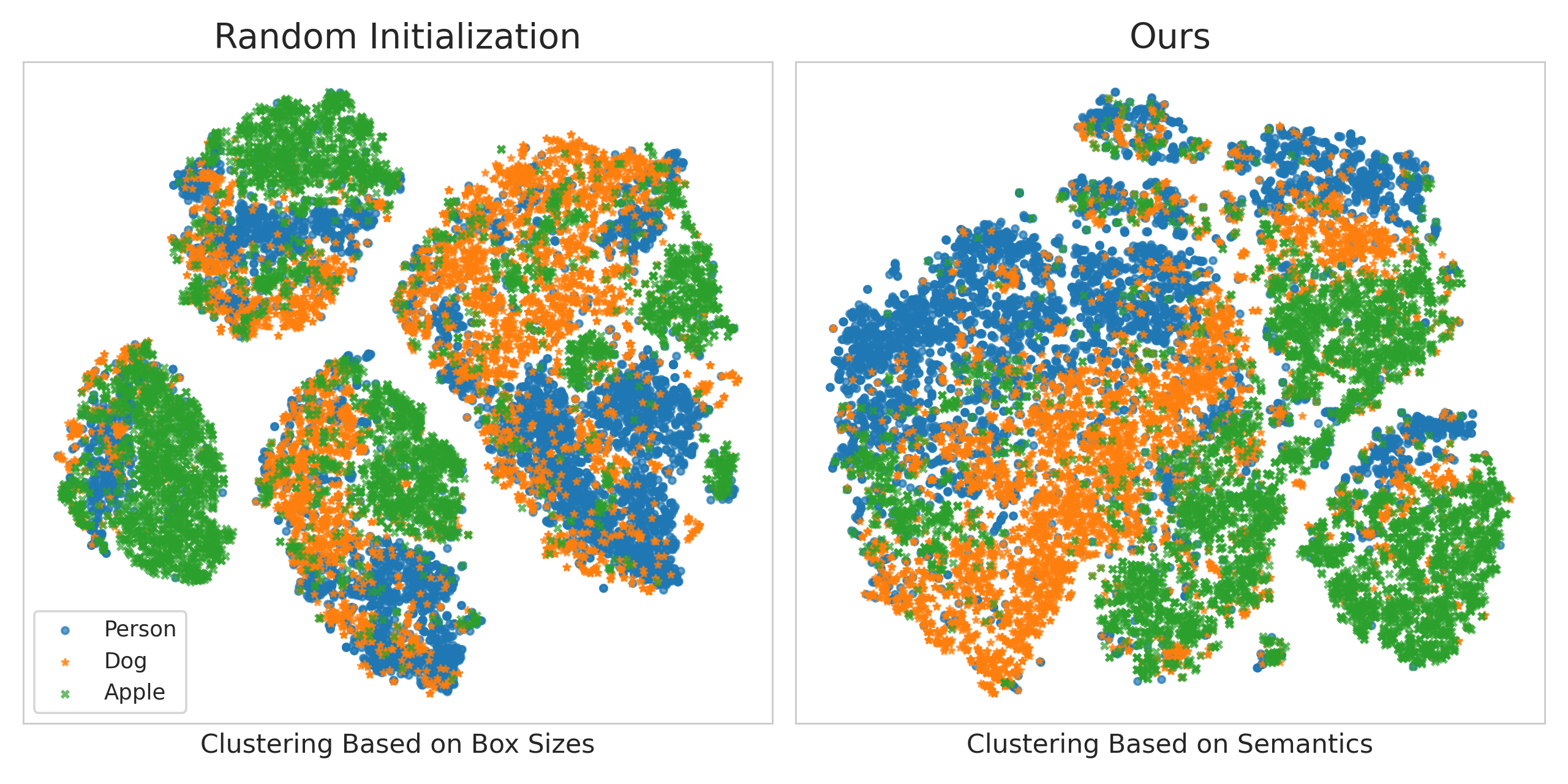}
    \caption{
    t-SNE visualization of ground truth annotations. AlignDet pre-training results in better class separation.
    }
    \label{fig:t-sne}
\end{figure}

\begin{table}[!]
\begin{center}
\resizebox{\linewidth}{!}{%
\begin{tabular}{c|c|cccccc}
\textbf{Image-domain}
& \textbf{Box-domain}
& \textbf{AP}
& \textbf{AP$_{50}$}
& \textbf{AP$_{75}$}
& \textbf{AP$_{s}$}
& \textbf{AP$_{m}$}
& \textbf{AP$_{l}$}
\\ \hline
ImageNet &  Random Initialization
&  37.9  &  58.2  &  41.1  & 22.0  & 41.4  & 48.6  \\ \hline
ImageNet &  ImageNet Subset
&  38.7  &  58.6  &  42.1  & 21.7  & 42.0  & 50.1  \\ \hline
ImageNet & COCO
&  \textbf{39.0}  &  \textbf{59.3}  &  \textbf{42.5}  & \textbf{22.3}  & \textbf{42.4}  &  \textbf{50.4}
\end{tabular}%
}
\end{center}
\caption{Ablation study on data discrepancy. Both the data characteristics and domains matter in pre-training.}
\vspace{-0.1cm}
\label{tab:ablation_data}
\end{table}
\begin{table}[!]
\begin{center}
\resizebox{\linewidth}{!}{%
\begin{tabular}{c|c|c|c|cccccc}
\textbf{Backbone} & \textbf{Neck} & \textbf{RPN} & \textbf{Head} & \textbf{AP} & \textbf{AP$_{50}$} & \textbf{AP$_{75}$} & \textbf{AP$_s$} & \textbf{AP$_m$} & \textbf{AP$_l$} \\ \hline
\cmark & \xmark & \xmark & \xmark  & 37.9 & 58.2 & 41.1 & 22.0 & 41.4 & 48.6 \\ \hline
\cmark & \cmark & \xmark & \xmark  & 38.4 & 58.5 & 41.8 & 22.2 & 41.7 & 50.0 \\ \hline
\cmark & \cmark & \cmark & \xmark  & 38.2 & 58.3 & 41.4 & 21.4 & 41.6 & 49.7 \\ \hline
\cmark & \cmark & \xmark & \cmark  & 38.8 & 58.8 & 42.5 & 21.8 & 42.2 & \textbf{50.6} \\ \hline
\cmark & \cmark & \cmark & \cmark  & \textbf{39.0}  &  \textbf{59.3}  &  \textbf{42.5}  & \textbf{22.3}  & \textbf{42.4}  &  50.4
\end{tabular}%
}
\end{center}
\caption{Ablation study on the model discrepancy. Each module benefits from pre-training and leads to improvements.}
\vspace{-0.1cm}
\label{tab:ablation_model}
\end{table}

\vspace{-0.2cm}

\paragraph{Task Discrepancy: Are detection-oriented pretext tasks helpful?}
As shown in Table~\ref{tab:ablation_task}, both the classification and regression pretext task can improve the detector accuracy, boosting the mAP from 37.9 to 38.7.
Furthermore, the improvement brought by the classification pretext task is mainly reflected in the AP$_{50}$, while the regression task comes from the AP$_{75}$.
It reveals that the classification pre-training brings higher prediction accuracy, while the regression pretext task helps to predict more accurate coordinates.

\vspace{-0.2cm}

\begin{figure}[t]\centering
    \includegraphics[width=1.0\linewidth]{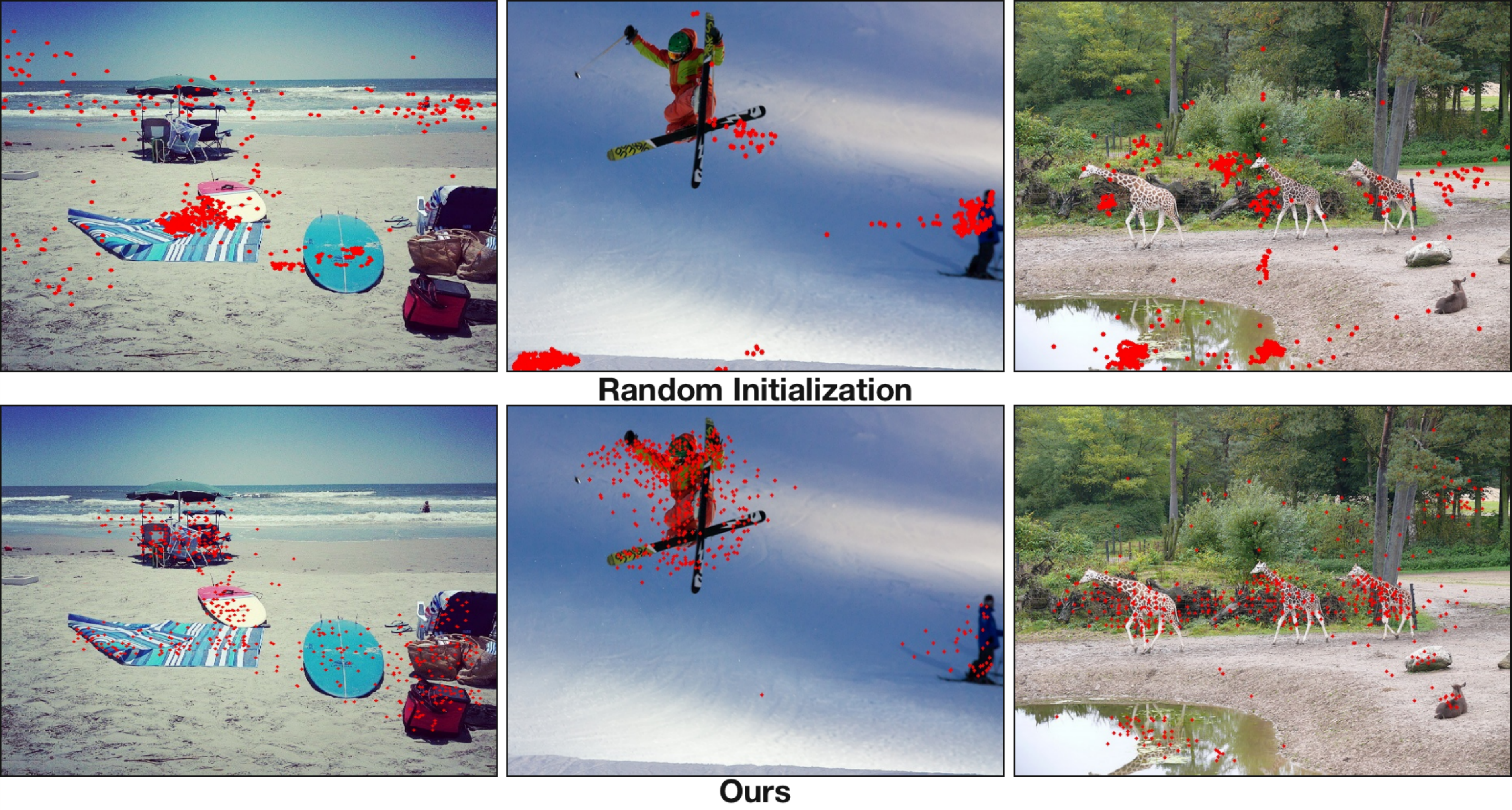}
    \caption{
    Visualization of predictions on COCO Val2017.
    }
    \label{fig:visualization}
\end{figure}

\paragraph{Effectiveness of Box-domain Pre-training.}
To visualize the effectiveness of box-domain pre-training, we depict t-SNE visualization~\cite{tsne} for features of 3 classes with each 5000 ground truth annotations. As illustrated in Figure~\ref{fig:t-sne}, the features from the randomly initialized contrastive head $f^{con}$ are clustered into four parts, with each cluster representing a feature level in the FPN.
It reveals that the random initialization head classifies each box based on the box size. 
However, AlignDet provides a good weight initialization for $f^{con}$, leading to better class separation. We also visualize the center points of predicted boxes from Faster R-CNN in Figure~\ref{fig:visualization}. AlignDet focuses on potential objects instead of messy pixels compared to the random initialization results without box-domain pre-training. There are more visualization results of other methods in the supplementary material.

\begin{table}[!]
\begin{center}
\resizebox{\linewidth}{!}{%
\begin{tabular}{c|cc|cccccc}
\multicolumn{1}{c|}{\textbf{Image-domain}} &
\multicolumn{2}{c|}{\textbf{Box-domain}} &
  \multirow{2}{*}{\textbf{AP}} &
  \multirow{2}{*}{\textbf{AP$_{50}$}} &
  \multirow{2}{*}{\textbf{AP$_{75}$}} &
  \multirow{2}{*}{\textbf{AP$_s$}} &
  \multirow{2}{*}{\textbf{AP$_m$}} &
  \multirow{2}{*}{\textbf{AP$_l$}} \\ \cline{1-3}
\multicolumn{1}{c|}{\textbf{Classification}} & \multicolumn{1}{c|}{\textbf{Classification}} & \textbf{Regression} &   &   &   &   &   &   \\ \hline
\cmark & \multicolumn{1}{c|}{\xmark} & \xmark & 37.9 & 58.2 & 41.1 & 22.0 & 41.4 & 48.6 \\ \hline
\cmark & \multicolumn{1}{c|}{\cmark} & \xmark & 38.7 & 58.8 & 42.1 & 22.4 & 41.9 & 50.2 \\ \hline
\cmark & \multicolumn{1}{c|}{\xmark} & \cmark & 38.7 & 58.6 & 42.4 & 22.1 & 41.8 & 50.2 \\ \hline
\cmark & \multicolumn{1}{c|}{\cmark} & \cmark & \textbf{39.0}  &  \textbf{59.3}  &  \textbf{42.5}  & \textbf{22.3}  & \textbf{42.4}  &  \textbf{50.4}
\end{tabular}%
}
\end{center}
\caption{Ablation study on task discrepancy. Both the classification and regression tasks are essential for AlignDet.}
\vspace{-0.1cm}
\label{tab:ablation_task}
\end{table}

\begin{figure*}[t!]\centering
    \includegraphics[width=1.0\linewidth]{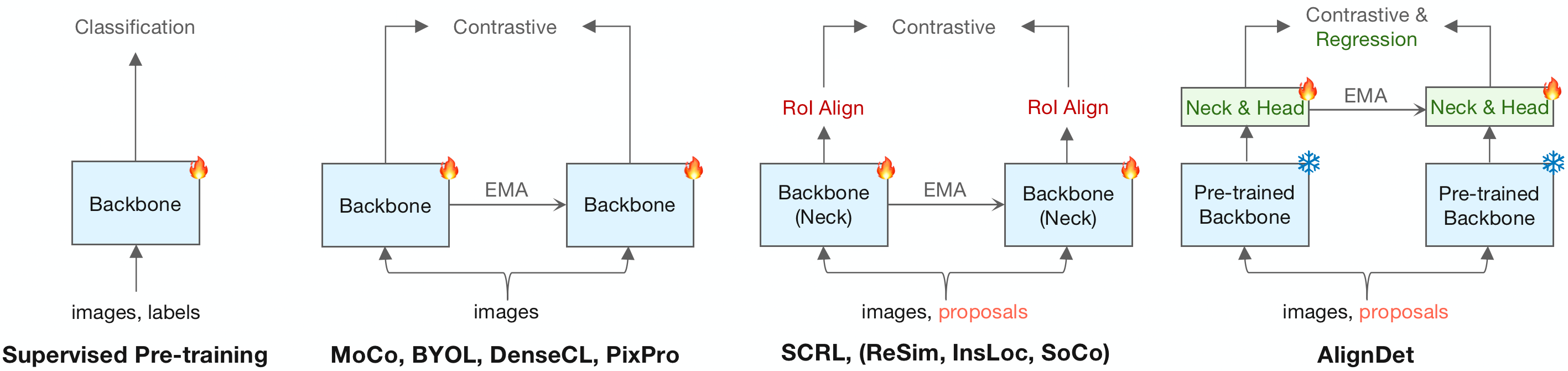}
    \caption{Compared with other pre-training methods, AlignDet achieves adequate and more efficient  pre-training for various detectors.}
    \label{fig:comparasion_supp}
\end{figure*}

\paragraph{Freeze Backbone in Box-domain Pre-training.}
Freezing a layer or all parameters in the backbone is a common practice in previous work~\cite{mask_rcnn,up_detr,detreg}. For example, Mask R-CNN freezes the first layer of ResNet when training on COCO, and UP-DETR and DETReg freeze the entire backbone during pre-training. Although there is no explanation for this approach in previous works, it significantly degrades the performance~\cite{up_detr,detreg}.  Here we offer some conjectures as to why the backbone weights should be fixed during the box-domain pre-training stage, demonstrating that the frozen backbone leads to better performance. 

The possible reason is that the amount of parameters is large for the detection model, while the data is relatively small, and the label is noisy. If we train the entire detection model with small and noisy labels, the detector may over-fit to these noise proposals, rather than learning semantic and position priors of objects.
Therefore, in the box-domain pre-training, we freeze all parameters in the backbone.  We also constructed experiments on Table~\ref{tab:frozen_backbone} to demonstrate that fixing the backbone can bring more benefits.

\vspace{-0.5cm}
\paragraph{Ablation on Training Complexity.}
As shown in Table~\ref{tab:ablation}, the box-domain pre-training introduces only a small computational overhead compared to the image-domain backbone pre-training (\eg, MoCo v2 here), yet achieves significant and wide-ranging improvements as shown in the paper, the training time is calculated on 8 NVIDIA A100 GPUs (80G). Please note we simply load the pre-trained backbone weights in practice, and both the selective search and our AlignDet are easy to implement.

\begin{table}[]
\resizebox{\columnwidth}{!}{%
\begin{tabular}{c|c|c|c|c|c|c}
\textbf{Frozen Backbone} & \textbf{AP} & \textbf{AP$_{50}$} & \textbf{AP$_{75}$} & \textbf{AP$_{s}$} & \textbf{AP$_{m}$} & \textbf{AP$_{l}$} \\ \hline
\xmark    & 35.5    & 53.7      & 37.7      & 18.5     & 38.9     & 47.0   \\ \hline
\cmark    & \textbf{37.3}    & \textbf{56.6}      & \textbf{40.1}      & \textbf{21.0}     & \textbf{40.9}     & \textbf{49.8}   \\
       
\end{tabular}%
}
\caption{Ablation study on weather freezes the backbone during the box-domain pre-training.}
\vspace{-0.2cm}
\label{tab:frozen_backbone}
\end{table}

\vspace{-0.2cm}
\section{Comparison with Other Methods}
\label{comparison}
Compared with box-level contrastive learning counterparts (\eg, InsLoc~\cite{insloc}, SoCo~\cite{soco}), AlignDet makes full use of existing pre-trained backbones to enable efficient detection-oriented pre-training.
We also show the pseudo-code of SoCo and AlignDet pre-training in the Supplement Material for a more intuitive comparison. In order to better understand the differences with other pre-training methods, we demonstrate the core pipelines of these methods in Figure~\ref{fig:comparasion_supp}.
Compared with existing pre-training methods, AlignDet achieves adequate pre-training for all modules by introducing classification and regression knowledge, thus comprehensively addressing data, model, and task discrepancies between pre-training and fine-tuning.

\begin{table}[!]
\resizebox{\columnwidth}{!}{%
\begin{tabular}{c|c|c|c|c}
\textbf{Training Stage} & \textbf{Parameters} & \textbf{Dataset} & \textbf{Epochs} & \textbf{Training Time} \\ \hline
Image-domain Pre-training & 23.23 M  & ImageNet &  800  & 77.1 h \\ \hline
Box-domain Pre-training   & 8.79 M   & COCO     &  12   & 5.6 h  \\ \hline
Fine-tuning  & 32.02 M  & COCO     &  12   & 4.4 h  \\
\end{tabular}%
}
\caption{Ablation on training complexity for FCOS (ResNet-50)}
\vspace{-0.3cm}
\label{tab:ablation}
\end{table}

\section{Discussions}
\label{discussion}

\paragraph{How do pre-training and fine-tuning attain alignment?}
AlignDet achieves alignments because the box-domain pre-training can be considered as a type of detection fine-tuning, the only differences are freezing the backbone and replacing the classification with box contrastive learning.
To achieve \textit{data alignment}, we utilize the same multi-object dataset to maintain uniformity in data properties and domain.
The decoupled pre-training makes all modules can be well pre-trained to achieve \textit{model alignment}.
\textit{Task alignment} is promoted via the integration of both regression and classification prior knowledge within the box-domain pre-training.

\paragraph{Is Selective Search Burdensome in Pre-training?}
No. The generation of selective search proposals is conducted offline and performed only once before pre-training. In box-domain pre-training, we randomly select the generated proposals as inputs, and operations of proposals do not bring too much overhead, as shown in Table~\ref{tab:ablation}.

\paragraph{Complexity of AlignDet.}
Although AlignDet requires three stages of image-domain pre-training, box-domain pre-training, and final fine-tuning. But we want to emphasize that both image-domain pre-training and fine-tuning are standard paradigms in object detection, so AlignDet only introduces an additional box-domain pre-training phase.

\vspace{0.2cm}
\section{Conclusion}
\label{sec:conclusion}
In this paper, we point out that there are data, model, and task discrepancies between pre-training and fine-tuning in object detection and propose a unified framework namely AlignDet to address these issues. AlignDet learns both classification and regression knowledge and enables highly efficient pre-training for all modules. Notably, it is the first framework to facilitate complete unsupervised pre-training of various detectors. Extensive experiments demonstrate that AlignDet significantly improves performance across diverse protocols. We believe AlignDet provides valuable insights and opens new avenues for visual pre-training.

\clearpage
\newpage

{\small
\bibliographystyle{ieee_fullname}
\bibliography{egbib}

\begin{thebibliography}{10}\itemsep=-1pt

\bibitem{beit}
Hangbo Bao, Li Dong, Songhao Piao, and Furu Wei.
\newblock Beit: Bert pre-training of image transformers.
\newblock In {\em ICLR}, 2021.

\bibitem{detreg}
Amir Bar, Xin Wang, Vadim Kantorov, Colorado~J Reed, Roei Herzig, Gal Chechik,
  Anna Rohrbach, Trevor Darrell, and Amir Globerson.
\newblock Detreg: Unsupervised pretraining with region priors for object
  detection.
\newblock In {\em CVPR}, 2022.

\bibitem{bigdet}
Likun Cai, Zhi Zhang, Yi Zhu, Li Zhang, Mu Li, and Xiangyang Xue.
\newblock Bigdetection: A large-scale benchmark for improved object detector
  pre-training.
\newblock In {\em CVPR}, 2022.

\bibitem{detr}
Nicolas Carion, Francisco Massa, Gabriel Synnaeve, Nicolas Usunier, Alexander
  Kirillov, and Sergey Zagoruyko.
\newblock End-to-end object detection with transformers.
\newblock In {\em ECCV}, 2020.

\bibitem{deep_cluster}
Mathilde Caron, Piotr Bojanowski, Armand Joulin, and Matthijs Douze.
\newblock Deep clustering for unsupervised learning of visual features.
\newblock In {\em ECCV}, 2018.

\bibitem{swav}
Mathilde Caron, Ishan Misra, Julien Mairal, Priya Goyal, Piotr Bojanowski, and
  Armand Joulin.
\newblock Unsupervised learning of visual features by contrasting cluster
  assignments.
\newblock {\em NeurIPS}, 2020.

\bibitem{mmdet}
Kai Chen, Jiaqi Wang, Jiangmiao Pang, Yuhang Cao, Yu Xiong, Xiaoxiao Li,
  Shuyang Sun, Wansen Feng, Ziwei Liu, Jiarui Xu, et~al.
\newblock Mmdetection: Open mmlab detection toolbox and benchmark.
\newblock {\em arXiv preprint arXiv:1906.07155}, 2019.

\bibitem{simclr}
Ting Chen, Simon Kornblith, Mohammad Norouzi, and Geoffrey Hinton.
\newblock A simple framework for contrastive learning of visual
  representations.
\newblock In {\em ICML}, 2020.

\bibitem{aqtc}
Tom~Tongjia Chen, Hongshan Yu, Zhengeng Yang, Ming Li, Zechuan Li, Jingwen
  Wang, Wei Miao, Wei Sun, and Chen Chen.
\newblock First place solution to the cvpr'2023 aqtc challenge: A
  function-interaction centric approach with spatiotemporal visual-language
  alignment.
\newblock {\em arXiv preprint arXiv:2306.13380}, 2023.

\bibitem{mocov2}
Xinlei Chen, Haoqi Fan, Ross Girshick, and Kaiming He.
\newblock Improved baselines with momentum contrastive learning.
\newblock {\em arXiv preprint arXiv:2003.04297}, 2020.

\bibitem{simsiam}
Xinlei Chen and Kaiming He.
\newblock Exploring simple siamese representation learning.
\newblock In {\em CVPR}, 2021.

\bibitem{up_detr}
Zhigang Dai, Bolun Cai, Yugeng Lin, and Junying Chen.
\newblock Up-detr: Unsupervised pre-training for object detection with
  transformers.
\newblock In {\em CVPR}, 2021.

\bibitem{inadequately}
Andong Deng, Xingjian Li, Zhibing Li, Di Hu, Chengzhong Xu, and Dejing Dou.
\newblock Inadequately pre-trained models are better feature extractors.
\newblock {\em arXiv preprint arXiv:2203.04668}, 2022.

\bibitem{imagenet}
Jia Deng, Wei Dong, Richard Socher, Li-Jia Li, Kai Li, and Li Fei-Fei.
\newblock Imagenet: A large-scale hierarchical image database.
\newblock In {\em CVPR}, 2009.

\bibitem{pretask_context}
Carl Doersch, Abhinav Gupta, and Alexei~A Efros.
\newblock Unsupervised visual representation learning by context prediction.
\newblock In {\em ICCV}, 2015.

\bibitem{voc}
Mark Everingham, Luc Van~Gool, Christopher~KI Williams, John Winn, and Andrew
  Zisserman.
\newblock The pascal visual object classes (voc) challenge.
\newblock {\em IJCV}, 2010.

\bibitem{scp}
Golnaz Ghiasi, Yin Cui, Aravind Srinivas, Rui Qian, Tsung-Yi Lin, Ekin~D Cubuk,
  Quoc~V Le, and Barret Zoph.
\newblock Simple copy-paste is a strong data augmentation method for instance
  segmentation.
\newblock In {\em CVPR}, 2021.

\bibitem{pretask_rotation}
Spyros Gidaris, Praveer Singh, and Nikos Komodakis.
\newblock Unsupervised representation learning by predicting image rotations.
\newblock In {\em ICLR}, 2018.

\bibitem{fast_rcnn}
Ross Girshick.
\newblock Fast r-cnn.
\newblock In {\em ICCV}, 2015.

\bibitem{rcnn}
Ross Girshick, Jeff Donahue, Trevor Darrell, and Jitendra Malik.
\newblock Rich feature hierarchies for accurate object detection and semantic
  segmentation.
\newblock In {\em CVPR}, 2014.

\bibitem{relu}
Xavier Glorot, Antoine Bordes, and Yoshua Bengio.
\newblock Deep sparse rectifier neural networks.
\newblock In {\em Aistats}, 2011.

\bibitem{byol}
Jean-Bastien Grill, Florian Strub, Florent Altch{\'e}, Corentin Tallec, Pierre
  Richemond, Elena Buchatskaya, Carl Doersch, Bernardo Avila~Pires, Zhaohan
  Guo, Mohammad Gheshlaghi~Azar, et~al.
\newblock Bootstrap your own latent-a new approach to self-supervised learning.
\newblock {\em NeurIPS}, 2020.

\bibitem{mae}
Kaiming He, Xinlei Chen, Saining Xie, Yanghao Li, Piotr Doll{\'a}r, and Ross
  Girshick.
\newblock Masked autoencoders are scalable vision learners.
\newblock In {\em CVPR}, 2022.

\bibitem{moco}
Kaiming He, Haoqi Fan, Yuxin Wu, Saining Xie, and Ross Girshick.
\newblock Momentum contrast for unsupervised visual representation learning.
\newblock In {\em CVPR}, 2020.

\bibitem{rethinking}
Kaiming He, Ross Girshick, and Piotr Doll{\'a}r.
\newblock Rethinking imagenet pre-training.
\newblock In {\em ICCV}, 2019.

\bibitem{mask_rcnn}
Kaiming He, Georgia Gkioxari, Piotr Doll{\'a}r, and Ross Girshick.
\newblock Mask r-cnn.
\newblock In {\em ICCV}, 2017.

\bibitem{resnet}
Kaiming He, Xiangyu Zhang, Shaoqing Ren, and Jian Sun.
\newblock Deep residual learning for image recognition.
\newblock In {\em CVPR}, 2016.

\bibitem{bn}
Sergey Ioffe and Christian Szegedy.
\newblock Batch normalization: Accelerating deep network training by reducing
  internal covariate shift.
\newblock In {\em ICML}, 2015.

\bibitem{openimages}
Alina Kuznetsova, Hassan Rom, Neil Alldrin, Jasper Uijlings, Ivan Krasin, Jordi
  Pont-Tuset, Shahab Kamali, Stefan Popov, Matteo Malloci, Alexander
  Kolesnikov, et~al.
\newblock The open images dataset v4: Unified image classification, object
  detection, and visual relationship detection at scale.
\newblock {\em IJCV}, 2020.

\bibitem{pretask_colorization}
Gustav Larsson, Michael Maire, and Gregory Shakhnarovich.
\newblock Colorization as a proxy task for visual understanding.
\newblock In {\em CVPR}, 2017.

\bibitem{cornernet}
Hei Law and Jia Deng.
\newblock Cornernet: Detecting objects as paired keypoints.
\newblock {\em IJCV}, 2019.

\bibitem{ppt}
Ming Li, Jie Wu, Jinhang Cai, Jie Qin, Yuxi Ren, Xuefeng Xiao, Min Zheng, Rui
  Wang, and Xin Pan.
\newblock Parallel pre-trained transformers (ppt) for synthetic data-based
  instance segmentation.
\newblock {\em arXiv preprint arXiv:2206.10845}, 2022.

\bibitem{vitdet}
Yanghao Li, Hanzi Mao, Ross Girshick, and Kaiming He.
\newblock Exploring plain vision transformer backbones for object detection.
\newblock {\em arXiv preprint arXiv:2203.16527}, 2022.

\bibitem{cbnet}
Tingting Liang, Xiaojie Chu, Yudong Liu, Yongtao Wang, Zhi Tang, Wei Chu,
  Jingdong Chen, and Haibin Ling.
\newblock Cbnet: A composite backbone network architecture for object
  detection.
\newblock {\em TIP}, 2022.

\bibitem{fpn}
Tsung-Yi Lin, Piotr Doll{\'a}r, Ross Girshick, Kaiming He, Bharath Hariharan,
  and Serge Belongie.
\newblock Feature pyramid networks for object detection.
\newblock In {\em CVPR}, 2017.

\bibitem{retinanet}
Tsung-Yi Lin, Priya Goyal, Ross Girshick, Kaiming He, and Piotr Doll{\'a}r.
\newblock Focal loss for dense object detection.
\newblock In {\em ICCV}, 2017.

\bibitem{coco}
Tsung-Yi Lin, Michael Maire, Serge Belongie, James Hays, Pietro Perona, Deva
  Ramanan, Piotr Doll{\'a}r, and C~Lawrence Zitnick.
\newblock Microsoft coco: Common objects in context.
\newblock In {\em ECCV}, 2014.

\bibitem{self_emd}
Songtao Liu, Zeming Li, and Jian Sun.
\newblock Self-emd: Self-supervised object detection without imagenet.
\newblock {\em arXiv preprint arXiv:2011.13677}, 2020.

\bibitem{ssd}
W. Liu, Dragomir Anguelov, D. Erhan, Christian Szegedy, Scott~E. Reed,
  Cheng-Yang Fu, and Alexander~C. Berg.
\newblock Ssd: Single shot multibox detector.
\newblock In {\em ECCV}, 2016.

\bibitem{swin}
Ze Liu, Yutong Lin, Yue Cao, Han Hu, Yixuan Wei, Zheng Zhang, Stephen Lin, and
  Baining Guo.
\newblock Swin transformer: Hierarchical vision transformer using shifted
  windows.
\newblock In {\em ICCV}, 2021.

\bibitem{kmeans}
Stuart Lloyd.
\newblock Least squares quantization in pcm.
\newblock {\em IEEE transactions on information theory}, 1982.

\bibitem{kmeans++}
I MacQueen.
\newblock Some methods for classifiction and analysis of multivariate
  observations.
\newblock In {\em Proceedings 5th Berkeley Symposium on Mathematical Statistics
  Problems}, 1967.

\bibitem{gfm}
Matias Mendieta, Boran Han, Xingjian Shi, Yi Zhu, Chen Chen, and Mu Li.
\newblock Gfm: Building geospatial foundation models via continual pretraining.
\newblock {\em arXiv preprint arXiv:2302.04476}, 2023.

\bibitem{pytorch}
Adam Paszke, Sam Gross, Francisco Massa, Adam Lerer, James Bradbury, Gregory
  Chanan, Trevor Killeen, Zeming Lin, Natalia Gimelshein, Luca Antiga, et~al.
\newblock Pytorch: An imperative style, high-performance deep learning library.
\newblock {\em NeurIPS}, 2019.

\bibitem{pretask_inpainting}
Deepak Pathak, Philipp Krahenbuhl, Jeff Donahue, Trevor Darrell, and Alexei~A
  Efros.
\newblock Context encoders: Feature learning by inpainting.
\newblock In {\em CVPR}, 2016.

\bibitem{qin2022multi}
Jie Qin, Jie Wu, Ming Li, Xuefeng Xiao, Min Zheng, and Xingang Wang.
\newblock Multi-granularity distillation scheme towards lightweight
  semi-supervised semantic segmentation.
\newblock In {\em ECCV}, 2022.

\bibitem{qin2023freeseg}
Jie Qin, Jie Wu, Pengxiang Yan, Ming Li, Ren Yuxi, Xuefeng Xiao, Yitong Wang,
  Rui Wang, Shilei Wen, Xin Pan, et~al.
\newblock Freeseg: Unified, universal and open-vocabulary image segmentation.
\newblock In {\em CVPR}, 2023.

\bibitem{selfsup_helps_selfsup}
Colorado~J Reed, Xiangyu Yue, Ani Nrusimha, Sayna Ebrahimi, Vivek Vijaykumar,
  Richard Mao, Bo Li, Shanghang Zhang, Devin Guillory, Sean Metzger, et~al.
\newblock Self-supervised pretraining improves self-supervised pretraining.
\newblock In {\em WACV}, 2022.

\bibitem{faster_rcnn}
Shaoqing Ren, Kaiming He, Ross Girshick, and Jian Sun.
\newblock Faster r-cnn: Towards real-time object detection with region proposal
  networks.
\newblock {\em NeurIPS}, 2015.

\bibitem{scrl}
Byungseok Roh, Wuhyun Shin, Ildoo Kim, and Sungwoong Kim.
\newblock Spatially consistent representation learning.
\newblock In {\em CVPR}, 2021.

\bibitem{mbv2}
Mark Sandler, Andrew Howard, Menglong Zhu, Andrey Zhmoginov, and Liang-Chieh
  Chen.
\newblock Mobilenetv2: Inverted residuals and linear bottlenecks.
\newblock In {\em CVPR}, 2018.

\bibitem{obj365}
Shuai Shao, Zeming Li, Tianyuan Zhang, Chao Peng, Gang Yu, Xiangyu Zhang, Jing
  Li, and Jian Sun.
\newblock Objects365: A large-scale, high-quality dataset for object detection.
\newblock In {\em ICCV}, 2019.

\bibitem{fcos}
Zhi Tian, Chunhua Shen, Hao Chen, and Tong He.
\newblock Fcos: Fully convolutional one-stage object detection.
\newblock In {\em ICCV}, 2019.

\bibitem{selective_search}
Jasper~RR Uijlings, Koen~EA Van De~Sande, Theo Gevers, and Arnold~WM Smeulders.
\newblock Selective search for object recognition.
\newblock {\em IJCV}, 2013.

\bibitem{tsne}
Laurens Van~der Maaten and Geoffrey Hinton.
\newblock Visualizing data using t-sne.
\newblock {\em JMLR}, 2008.

\bibitem{tau}
Feng Wang and Huaping Liu.
\newblock Understanding the behaviour of contrastive loss.
\newblock In {\em CVPR}, 2021.

\bibitem{densecl}
Xinlong Wang, Rufeng Zhang, Chunhua Shen, Tao Kong, and Lei Li.
\newblock Dense contrastive learning for self-supervised visual pre-training.
\newblock In {\em CVPR}, 2021.

\bibitem{soco}
Fangyun Wei, Yue Gao, Zhirong Wu, Han Hu, and Stephen Lin.
\newblock Aligning pretraining for detection via object-level contrastive
  learning.
\newblock {\em NeurIPS}, 2021.

\bibitem{slotcon}
Xin Wen, Bingchen Zhao, Anlin Zheng, Xiangyu Zhang, and Xiaojuan Qi.
\newblock Self-supervised visual representation learning with semantic
  grouping.
\newblock {\em NeurIPS}, 2022.

\bibitem{wu2020fine}
Jie Wu, Tianshui Chen, Hefeng Wu, Zhi Yang, Guangchun Luo, and Liang Lin.
\newblock Fine-grained image captioning with global-local discriminative
  objective.
\newblock {\em TMM}, 2020.

\bibitem{wu2020tree}
Jie Wu, Guanbin Li, Si Liu, and Liang Lin.
\newblock Tree-structured policy based progressive reinforcement learning for
  temporally language grounding in video.
\newblock In {\em AAAI}, 2020.

\bibitem{detco}
Enze Xie, Jian Ding, Wenhai Wang, Xiaohang Zhan, Hang Xu, Peize Sun, Zhenguo
  Li, and Ping Luo.
\newblock Detco: Unsupervised contrastive learning for object detection.
\newblock In {\em CVPR}, 2021.

\bibitem{pixpro}
Zhenda Xie, Yutong Lin, Zheng Zhang, Yue Cao, Stephen Lin, and Han Hu.
\newblock Propagate yourself: Exploring pixel-level consistency for
  unsupervised visual representation learning.
\newblock In {\em CVPR}, 2021.

\bibitem{simmim}
Zhenda Xie, Zheng Zhang, Yue Cao, Yutong Lin, Jianmin Bao, Zhuliang Yao, Qi
  Dai, and Han Hu.
\newblock Simmim: A simple framework for masked image modeling.
\newblock In {\em CVPR}, 2022.

\bibitem{soft_teacher}
Mengde Xu, Zheng Zhang, Han Hu, Jianfeng Wang, Lijuan Wang, Fangyun Wei, Xiang
  Bai, and Zicheng Liu.
\newblock End-to-end semi-supervised object detection with soft teacher.
\newblock In {\em ICCV}, 2021.

\bibitem{insloc}
Ceyuan Yang, Zhirong Wu, Bolei Zhou, and Stephen Lin.
\newblock Instance localization for self-supervised detection pretraining.
\newblock In {\em CVPR}, 2021.

\bibitem{reppoints}
Ze Yang, Shaohui Liu, Han Hu, Liwei Wang, and Stephen Lin.
\newblock Reppoints: Point set representation for object detection.
\newblock {\em ICCV}, 2019.

\bibitem{sam_detr}
Gongjie Zhang, Zhipeng Luo, Yingchen Yu, Kaiwen Cui, and Shijian Lu.
\newblock Accelerating detr convergence via semantic-aligned matching.
\newblock In {\em CVPR}, 2022.

\bibitem{dino}
Hao Zhang, Feng Li, Shilong Liu, Lei Zhang, Hang Su, Jun Zhu, Lionel Ni, and
  Heung-Yeung Shum.
\newblock Dino: Detr with improved denoising anchor boxes for end-to-end object
  detection.
\newblock In {\em ICLR}, 2022.

\bibitem{densesiam}
Wenwei Zhang, Jiangmiao Pang, Kai Chen, and Chen~Change Loy.
\newblock Dense siamese network.
\newblock {\em ECCV}, 2022.

\bibitem{ibot}
Jinghao Zhou, Chen Wei, Huiyu Wang, Wei Shen, Cihang Xie, Alan Yuille, and Tao
  Kong.
\newblock Image bert pre-training with online tokenizer.
\newblock In {\em ICLR}, 2021.

\bibitem{centernet}
Xingyi Zhou, Dequan Wang, and Philipp Kr{\"a}henb{\"u}hl.
\newblock Objects as points.
\newblock {\em arXiv preprint arXiv:1904.07850}, 2019.

\bibitem{extremenet}
Xingyi Zhou, Jiacheng Zhuo, and Philipp Krahenbuhl.
\newblock Bottom-up object detection by grouping extreme and center points.
\newblock In {\em CVPR}, 2019.

\bibitem{deformable_detr}
Xizhou Zhu, Weijie Su, Lewei Lu, Bin Li, Xiaogang Wang, and Jifeng Dai.
\newblock Deformable detr: Deformable transformers for end-to-end object
  detection.
\newblock In {\em ICLR}, 2020.

\end{thebibliography}
}

\clearpage
\newpage

\setcounter{section}{0}
\setcounter{algorithm}{0}
\renewcommand{\thesection}{\Alph{section}}
\renewcommand{\thealgorithm}{\Roman{algorithm}}
\section{Overview of Supplementary Material}
The supplementary material is organized into the following sections:

\begin{itemize}
    \item Section~\ref{details}: Implementation details for all experiments.
    \item Section~\ref{exps}: More experiments and analysis.
    \item Section~\ref{visualization}: Visualization of the selective search proposals and the effectiveness of AlignDet.
    \item Section~\ref{impact_and_limitation}: Broader impact and limitation.
\end{itemize}

\section{Implementation Details}
\label{details}
\subsection{General Settings}
\paragraph{Pre-training.}
All the hyper-parameters for box-domain pre-training follow the original fine-tuning settings except the prediction sampling procedure. For example, the learning rate is 2e-4 and weight decay is 1e-4 for Mask R-CNN~\cite{mask_rcnn} when fine-tuning on COCO~\cite{coco} with ImageNet~\cite{imagenet} pre-trained backbone. Hence in the box-domain pre-training stage, we set the same learning rate and weight decay to pre-train the modules out of the backbone. In terms of sampling predicted boxes, we select as many positive samples (predicted boxes that correspond to a ground truth proposal instead of background) as possible to expand the data for box-level contrastive learning. All methods apply the same pre-training data augmentation, which has been described in Section 4.1 of the main paper. All the experiments are pre-trained on the COCO train 2017 dataset with 12 epochs (1x), except 50 epochs for DETR~\cite{detr}. During the pre-training stage, most experiments can be finished with 8 V100 GPUs (32 GB), which is efficient since we only train other modules out of the backbone (\ie, neck and head).

\paragraph{Fine-tuning.}
During the fine-tuning stage, we use SyncBN~\cite{bn} to calibrate magnitudes for pre-trained models following MoCo~\cite{moco}. For the experiments with supervised pre-trained ResNet~\cite{resnet}, we follow the default setting in mmdetection~\cite{mmdet} to freeze the first layer of ResNet, and fine-tuning the other parameters under standard data augmentation with single-scale training. For the experiments with self-supervised backbones, we fine-tune all layers end-to-end with multi-scale training, and SyncBN is used across all layers, including the newly initialized batch normalization layers. For experiments with MobileNetv2~\cite{mbv2} and Swin Transformers~\cite{swin}, we follow the default training strategy defined in mmdetection.
For the VOC~\cite{voc} fine-tuning, we train 12k iterations to avoid over-fitting, and the learning rate is divided by 10 at $\frac{3}{4}$ and $\frac{11}{12}$ of total training time.

Since AlignDet pre-trains all modules in the detector and not just the backbone, we need to adjust the fine-tuned hyper-parameters to better transfer the pre-trained weights. Thanks to the experience of previous work~\cite{soco,obj365,vitdet}, adjusting the learning rate and weight decay is a good practice. The main principle of hyper-parameter adjustment in the fine-tuning stage is to increase the learning rate while reducing weight decay. The most common setting is to increase the learning rate by 1.5 times and reduce the weight decay to half of the original value. The specific values of different methods and experiments are listed in detail in each subsequent paragraph.

\subsection{FCOS}
FCOS~\cite{fcos} is a single-stage, point-based detector. The learning rate and weight decay are 0.1, 1e-4 for AlignDet pre-training and 0.15, 5e-5 for fine-tuning, respectively. The maximum number of sampled predicted boxes for the box-domain pre-training is 2048. Other hyper-parameters are set to the default values in mmdetection.

\subsection{RetinaNet}
RetinaNet~\cite{retinanet} is a single-stage, anchor-based detector. The learning rate and weight decay are 0.1, 1e-4 for AlignDet pre-training and 0.15, 5e-5 for fine-tuning, respectively. The maximum number of sampled predicted boxes for the box-domain pre-training is 2048. Other hyper-parameters are set to the default values in mmdetection.

\subsection{Faster R-CNN \& Mask R-CNN}
Faster R-CNN~\cite{fast_rcnn} and Mask R-CNN~\cite{mask_rcnn} are two-stage, anchor-based detectors. Here Faster R-CNN uses the RoI Align~\cite{mask_rcnn} operation. The maximum number of sampled predicted boxes for the box-domain pre-training is 4096. All the experiments including baseline results are re-implemented with the \textit{4conv1fc} RoI head for a fair comparison, following previous work~\cite{rethinking,moco}. For Faster R-CNN, we fine-tune with only object detection annotations, and for Mask R-CNN, we fine-tune with both object detection and instance segmentation annotations.

Specifically, for the supervised pre-trained MobileNet v2 and ResNet backbones, the learning rate and weight decay are 0.2, 1e-4 for AlignDet pre-training and 0.3, 5e-5 for fine-tuning, respectively.  In our experiments, the weight decay should be smaller for the self-supervised ResNet-50 backbones, thus we set 5e-6 for PixPro~\cite{pixpro} and MoCo v2~\cite{mocov2}, and the learning rate is the same as pre-training, \emph{i.e.}, 0.02. For SwAV~\cite{swav} pre-trained backbone, the fine-tuning learning rate is 3e-2, weight decay is 5e-6, and warmup iterations are 1000.
For Swin Transformer backbones, the learning rate is 1e-4 and weight decay is 5e-2 for AlignDet pre-training. During the fine-tuning stage, the learning rate is 1e-4 and weight decay is 2e-2.

\begin{algorithm}[t]
\caption{SoCo Pseudocode, PyTorch-like}
\label{code:soco}
\definecolor{codeblue}{rgb}{0.25,0.5,0.5}
\definecolor{codekw}{rgb}{0.85, 0.18, 0.50}
\lstset{
  backgroundcolor=\color{white},
  basicstyle=\fontsize{6.5pt}{6.5pt}\ttfamily\selectfont,
  columns=fullflexible,
  breaklines=true,
  captionpos=b,
  commentstyle=\fontsize{6.5pt}{6.5pt}\color{codeblue},
  keywordstyle=\fontsize{6.5pt}{6.5pt}\color{codekw},
}
\begin{lstlisting}[language=python]
# x: input images
# p: selective search proposals
# aug: independent random augmentation

for x, p in data_loader:
    (x1, p1), (x2, p2) = aug(x, p), aug(x, p) # augmentation
    x1, x2 = backbone_q(x1), backbone_k(x2) # updated
    x1, x2 = neck_q(z1), neck_k(z2)  # feature pyramid

    # proposals as final bboxes
    b1, b2 = p1, p2
    z1, z2 = roi_align(x1, p1), roi_align(x2, p2)
    z1, z2 = g_q(z1), g_k(z2)  # feature projection

    L = loss_constrastive(z1, z2)  # contrastive loss
    ema_update(backbone_q, backbone_k, neck_q, neck_k)
\end{lstlisting}
\end{algorithm}

\subsection{DETR}
DETR~\cite{detr} is a single-stage, query-based detector. A key factor that leads to slow convergence is the complication in aligning object queries with target features in different feature embedding spaces~\cite{sam_detr}. However, in the self-supervised setting, it is difficult to achieve this alignment because we do not have accurate semantic labels. To alleviate this issue, UP-DETR~\cite{up_detr} initializes the query embedding with features extracted from cropped image patches. DETReg~\cite{detreg} predict the features of cropped image patches from the corresponding query embedding via $L_1$ loss. However, these approaches simply use foreground or background for bipartite matching under the unsupervised setting, lacking explicit semantic information for the label assignment. This paradigm leads to the mismatch between bipartite matching costs and loss calculation, which may cause unstable matching and affect the effectiveness of pre-training.

To address this challenge, we make a small modification to AlignDet. In addition to the common coordinate-based label assignment and contrastive learning, which is the same in other methods, we also introduce the category-based assignment and corresponding loss to pre-train DETR. Specially, we crop the selective search~\cite{selective_search} proposals from images and extract their features with supervised pre-trained backbones. Then we cluster the extracted features into 256 classes using the K-means algorithm~\cite{kmeans,kmeans++}, the cluster results are regarded as pseudo-semantic labels to perform extra label assignment and cross-entropy loss to pre-train DETR. This has the advantage of introducing explicit category information into bipartite matching, which aligns label assignment and loss calculation in DETR, leading to more stable matching results.
\textit{Note that only DETR uses the clustering results of the features as extra pseudo-labels for box-domain pre-training, since the label assignment of other methods in this paper does not require explicit semantic information but only coordinates.}

We use both the default supervised pre-trained ResNet-50~\cite{resnet} and the self-supervised pre-trained SwAV~\cite{swav} for the experiments. The learning rate is 2e-4 for a batch size of 64 during the box-domain pre-training stage, and the loss weights of contrastive loss and cross-entropy loss are 1.0. In the fine-tuning stage, the learning rate is 1e-4 for the batch size of 16, and we fine-tune all the parameters following previous work~\cite{up_detr,detreg}. Other hyper-parameters are set to the default values in mmdetection.

\subsection{SimMIM and CBNet v2}
To further verify the effectiveness of AlignDet, we conducted advanced experiments with mask image modeling pre-training method (SimMIM~\cite{simmim}) and SOTA detection algorithm (CBNet v2~\cite{cbnet}). We chose CBNet v2 because of its open source code and achieved SOTA performance without requiring additional training data (\eg training on Objects365~\cite{obj365}). However, since they do not open source the training code corresponding to the most powerful model, we use the officially released code, models, and configs to reproduce the results. 
More specifically, we use the \href{https://github.com/open-mmlab/mmdetection/blob/master/configs/common/ssj_270k_coco_instance.py}{large scale jittering}~\cite{scp} to fine-tune Mask R-CNN with 3x strategy (SimMIM pre-trained Swin-Large backbone), following the settings reported in the original paper. For CBNet v2, we use the \href{https://github.com/VDIGPKU/CBNetV2/blob/main/configs/cbnet/htc_cbv2_swin_large_patch4_window7_mstrain_400-1400_giou_4conv1f_adamw_1x_coco.py}{publicly released config}~\cite{cbnet} to reproduce the results. Both external links are existing implementations that follow original papers, not part of our submission.

\begin{algorithm}[t]
\caption{AlignDet Pseudocode, PyTorch-like}
\label{code:aligndet}
\definecolor{codeblue}{rgb}{0.25,0.5,0.5}
\definecolor{codekw}{rgb}{0.85, 0.18, 0.50}
\lstset{
  backgroundcolor=\color{white},
  basicstyle=\fontsize{6.5pt}{6.5pt}\ttfamily\selectfont,
  columns=fullflexible,
  breaklines=true,
  captionpos=b,
  commentstyle=\fontsize{6.5pt}{6.5pt}\color{codeblue},
  keywordstyle=\fontsize{6.5pt}{6.5pt}\color{codekw},
}
\begin{lstlisting}[language=python]
# x: input images
# p: selective search proposals
# aug: independent random augmentation

for x, p in data_loader:
    (x1, p1), (x2, p2) = aug(x, p), aug(x, p) # augmentation
    x1, x2 = backbone(x1), backbone(x2)  # frozen backbone
    x1, x2 = neck_q(x1), neck_k(x2)  # feature pyramid

    # proposals as pseudo labels, boxes are predicted
    b1, b2 = head_q.f_reg(x1, p1), head_k.f_reg(x2, p2)
    z1, z2 = head_q.f_con(x1, b1), head_k.f_con(x2, b2)
    z1, z2 = g_q(z1), g_k(z2)  # feature projection

    L = loss_con(z1, z2) + loss_reg(b1, b2, p1, p2)  # losses
    ema_update(neck_q, neck_k, head_q, head_k)
\end{lstlisting}
\end{algorithm}

\section{Further Analysis and Experiments}
\label{exps}

\subsection{Pre-training with Longer Epochs}
Pre-training the backbone for longer epochs does not necessarily lead to sustained performance improvements for downstream tasks, both for supervised~\cite{rethinking} and self-supervised pre-training methods~\cite{soco,pixpro,moco}.
Here we find similar results on AlignDet, that is, 12 epochs pre-training is enough for AlignDet, as shown in Table~\ref{tab:ablation_pretrain_epoch} with RetinaNet.
However, the pre-training for the backbone are usually hundreds of epochs.
A potential reason for this phenomenon is that the pre-training parameters of the two are significantly different. In most object detection models, the number of parameters of the backbone is much more than that of the neck and head modules, so backbone pre-training often requires longer pre-training epochs to learn meaningful representation. On the contrary, since the neck and head modules have relatively few parameters, they can be well-trained with fewer epochs. Thus a longer pre-training time may lead to over-fitting and will not bring additional improvements. In addition, detection datasets such as COCO~\cite{coco} are usually smaller than pre-training datasets (\eg, ImageNet~\cite{imagenet}), which may exacerbate this issue.

\begin{table}[]
\resizebox{\columnwidth}{!}{%
\begin{tabular}{c|c|c|c|c|c|c}
\textbf{Pre-training Schedule} & \textbf{AP} & \textbf{AP$_{50}$} & \textbf{AP$_{75}$} & \textbf{AP$_{s}$} & \textbf{AP$_{m}$} & \textbf{AP$_{l}$} \\ \hline
1x & \textbf{37.3} & \textbf{56.6} & \textbf{40.1} & \textbf{21.0} & \textbf{40.9} & \textbf{49.8}  \\ \hline
2x & 37.0 & 56.2 & 39.3 & 20.8 & 40.6 & 49.5 \\ \hline
3x & 37.0 & 56.1 & 39.5 & 20.4 & 40.6 & 48.7
\\ \bottomrule
\end{tabular}%
}
\caption{Ablation study on pre-training schedules. All the results are fine-tuned with 12 epochs (1x schedule).}
\label{tab:ablation_pretrain_epoch}
\end{table}

\section{Visualization}
\label{visualization}
\subsection{Selective Search Proposals}
\begin{figure*}[t!]\centering
    \includegraphics[width=0.84\linewidth]{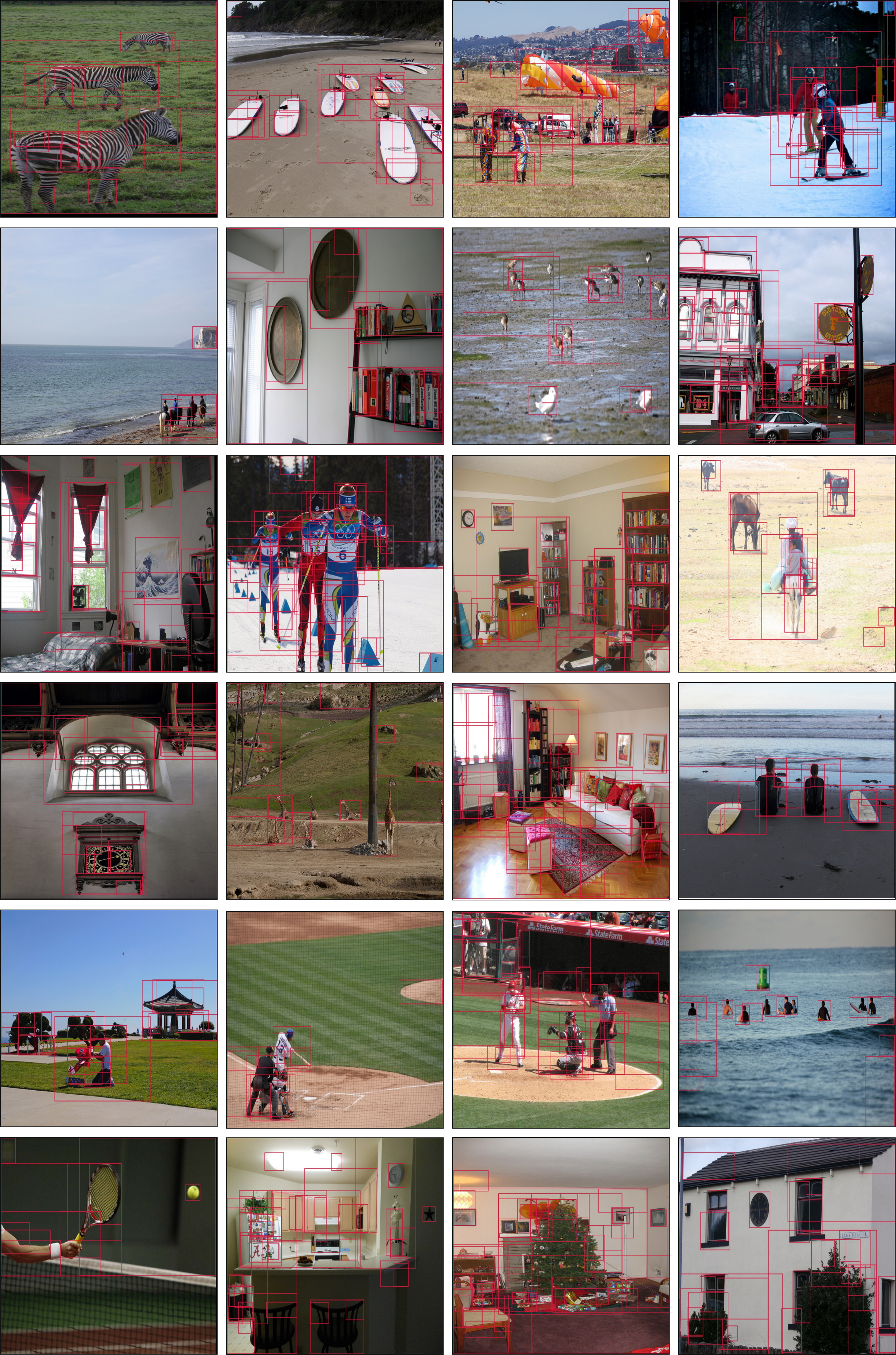}
    \caption{Selective search proposals on COCO train 2017 dataset.}
    \label{fig:selective_search_demo}
\end{figure*}

We use the same selective search code and filtering strategy as SoCo~\cite{soco} on the COCO train 2017 dataset, and apply non-maximum suppression (NMS) with a threshold of 0.5 at the end to remove redundant proposals. The images used for box-domain pre-training are shown in Figure~\ref{fig:selective_search_demo}.

\subsection{Effectiveness of Box-domain Pre-training}
Due to the significant differences in the design and mechanism of different detection methods, we need to design different visualization schemes to verify the effectiveness of our AlignDet under the unsupervised setting.

\begin{figure*}[t!]\centering
    \includegraphics[width=0.8\linewidth]{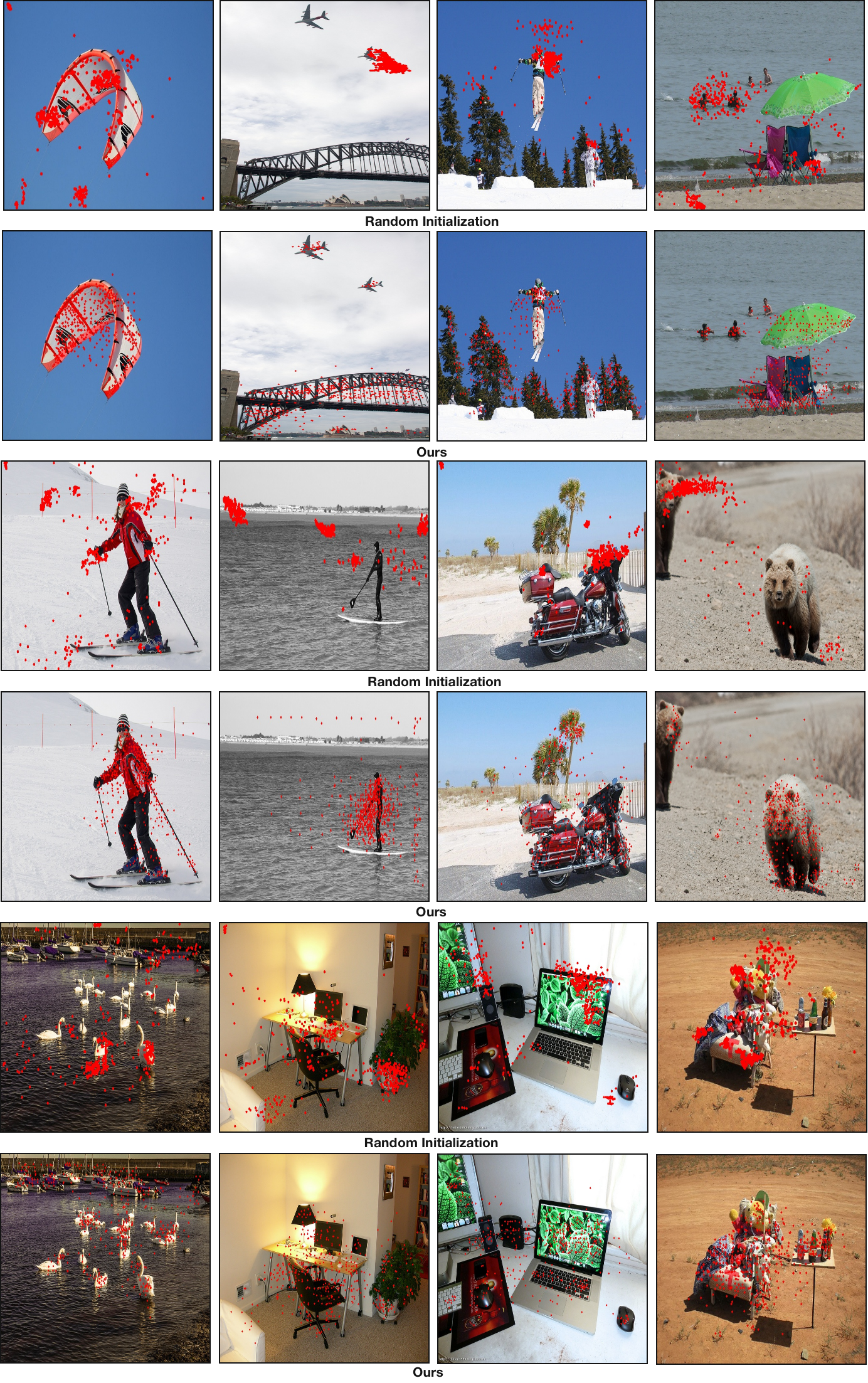}
    \vspace{-0.2cm}
    \caption{Visualization results of predictions on COCO Val 2017 with Faster/Mask R-CNN. Random Initialization denotes ImageNet pre-train, and ours means AlignDet pre-training.}
    \label{fig:faster_rcnn_demo}
\end{figure*}

\paragraph{Faster R-CNN \& Mask R-CNN.}
In this paper, the structural difference between Faster R-CNN and Mask R-CNN is only the presence or absence of a mask head, so they have the same prediction results and visualization for the detection task. In addition to the main paper, we also provide more visualizations here in Figure~\ref{fig:faster_rcnn_demo}. Specifically, we use RPN to determine which of the predicted boxes are foregrounds and feed them into the head to get the predicted box coordinates. We plot the centers of these boxes instead of rectangles for better visualization. AlignDet focuses on objects instead of messy pixels compared to the random initialization results without box-domain pre-training.

\begin{figure}[t!]\centering
    \includegraphics[width=1.0\linewidth]{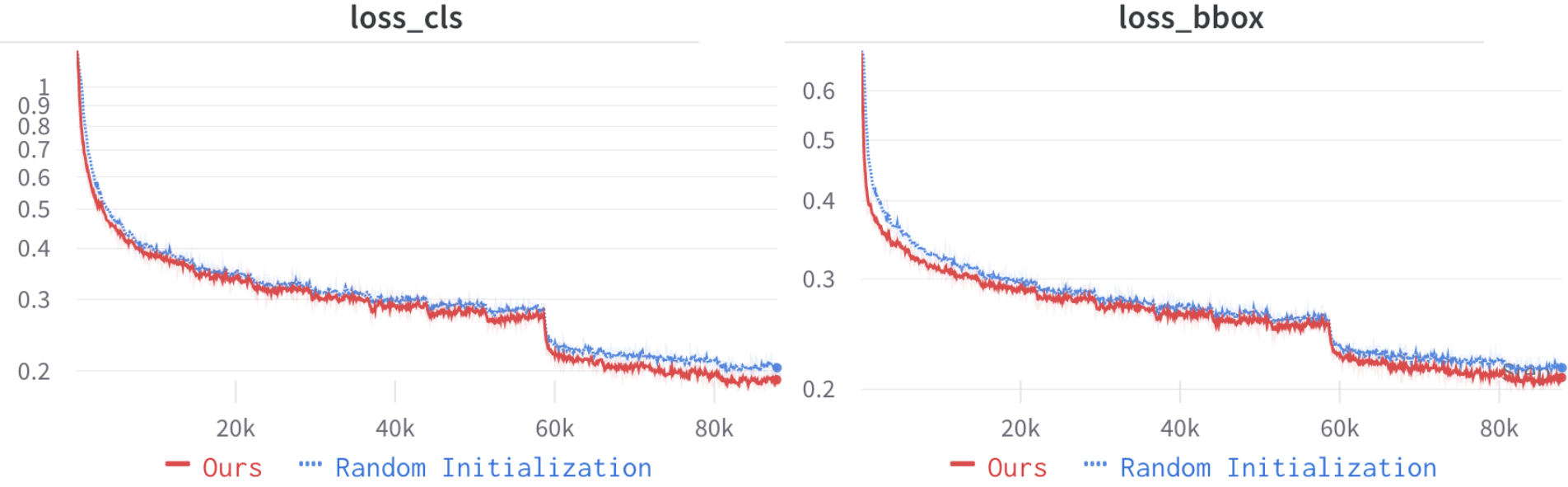}
    \caption{Fine-tuning losses of RetinaNet on COCO train 2017.}
    \label{fig:retinanet_loss}
\end{figure}

\paragraph{Other Methods.}
Unlike Faster R-CNN or Mask R-CNN, other methods do not have an RPN module, which means we cannot determine which predicted boxes are foreground and which are background during inference. To demonstrate the effectiveness of AlignDet, we show the training losses in Figure~\ref{fig:retinanet_loss} using RetinaNet as an example. AlignDet pre-training significantly accelerates the convergence of the model, with lower classification loss \textit{loss\_cls} and regression loss \textit{loss\_bbox} under the same training iterations.

\begin{figure*}[t!]\centering
    \includegraphics[width=0.8\linewidth]{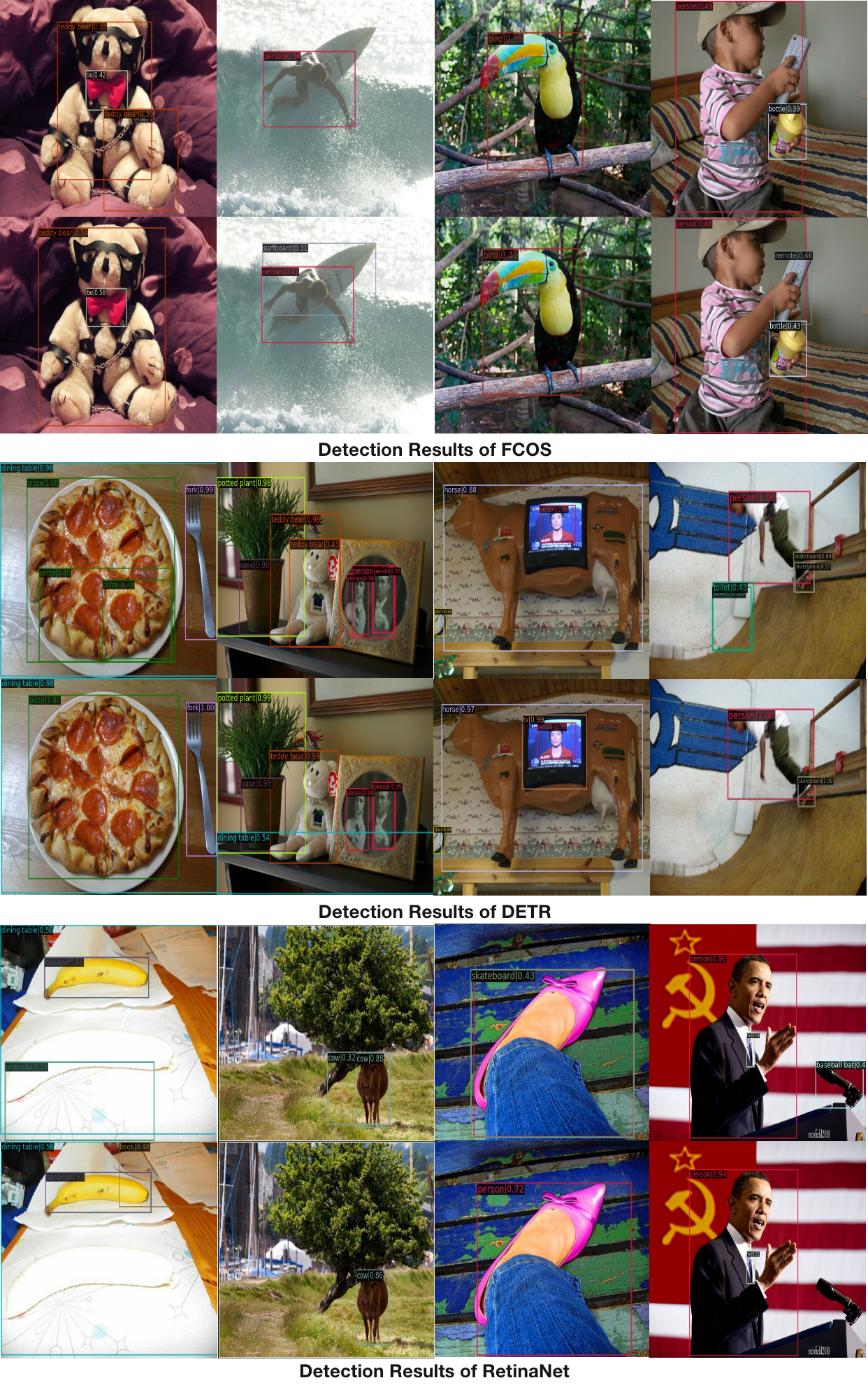}
    \vspace{-0.2cm}
    \caption{Detection results with different models on the public COCO Val 2017 dataset. For each scene, the upper images are the fine-tuning results without box-domain pre-training, and the lower images are the fine-tuning results after the box-domain pre-training.}
    
    \label{fig:demo}
\end{figure*}

In addition, we also show the fine-tuning results of different detection models with or without AlignDet pre-training in Figure~\ref{fig:demo} to further demonstrate the effectiveness of our AlignDet. AlignDet achieves more accurate classification and precise coordinate results than the random initialization results without box-domain pre-training.

\section{Broader Impact and Limitation}
\label{impact_and_limitation}
AlignDet represents a significant step forward in the development of unified and adequate unsupervised detection pre-training. Our approach enables the fully self-supervised pre-training of various object detection models, a milestone that was previously unattainable. Furthermore, the decoupled pre-training paradigm delivers highly efficient and effective pre-training, by separating the feature extraction from task-aware learning.\textbf{ The decoupled pre-training paradigm can be readily extended to other vision tasks, allowing the integration of general-purpose pre-trained backbones with task-aware pre-trained necks and heads, which opens a door for solving the discrepancies between general pre-training and various downstream tasks.}

However, the dependence on selective search proposals in this paper may represent a potential limitation, we view it as a direction for future research. Overall, our work advances the state-of-the-art unsupervised detection pre-training and offers significant potential for improving the performance of object detection.

\clearpage

\end{document}